\documentclass[letterpaper]{article} %
\usepackage{aaai25}  %
\usepackage{times}  %
\usepackage{helvet}  %
\usepackage{courier}  %
\usepackage[hyphens]{url}  %
\usepackage{graphicx} %
\usepackage{natbib}  %
\usepackage{caption} %

\usepackage[linesnumbered,ruled,vlined]{algorithm2e}
\usepackage{booktabs}       %
\usepackage{amsfonts}       %
\usepackage{xcolor}         %
\usepackage{subfigure}
\usepackage{amsmath}
\usepackage{amssymb}
\usepackage{mathtools}
\usepackage{amsthm}
\usepackage[capitalize,noabbrev]{cleveref}
\usepackage{multirow}
\usepackage{makecell}

\title{Structural Pruning via Spatial-aware Information Redundancy \\for Semantic Segmentation}
\author{
    Dongyue Wu\textsuperscript{\rm 1}, Zilin Guo\textsuperscript{\rm 1}, Li Yu\textsuperscript{\rm 2}, Nong Sang\textsuperscript{\rm 1}, Changxin Gao\textsuperscript{\rm 1}\thanks{Corresponding author.}\\
}
\affiliations{
    \textsuperscript{\rm 1}National Key Laboratory of Multispectral Information Intelligent Processing Technology, School of Artifcial Intelligence and Automation, Huazhong University of Science and Technology\\
    \textsuperscript{\rm 2}School of Electronic Information and Communications, Huazhong University of Science and Technology\\
    \{dongyue\_wu, zilin\_guo, hustlyu, nsang, cgao\}@hust.edu.cn
}

\usepackage{bibentry}

\begin{document}

\maketitle

\begin{abstract}
In recent years, semantic segmentation has flourished in various applications.
However, the high computational cost remains a significant challenge that hinders its further adoption. 
The filter pruning method for structured network slimming offers a direct and effective solution for the reduction of segmentation networks. 
Nevertheless, we argue that most existing pruning methods, originally designed for image classification, overlook the fact that segmentation is a location-sensitive task, which consequently leads to their suboptimal performance when applied to segmentation networks.
To address this issue, this paper proposes a novel approach, denoted as Spatial-aware Information Redundancy Filter Pruning~(SIRFP), which aims to reduce feature redundancy between channels. 
First, we formulate the pruning process as a maximum edge weight clique problem~(MEWCP) in graph theory, thereby minimizing the redundancy among the remaining features after pruning. 
Within this framework, we introduce a spatial-aware redundancy metric based on feature maps, thus endowing the pruning process with location sensitivity to better adapt to pruning segmentation networks. 
Additionally, based on the MEWCP, we propose a low computational complexity greedy strategy to solve this NP-hard problem, making it feasible and efficient for structured pruning. 
To validate the effectiveness of our method, we conducted extensive comparative experiments on various challenging datasets.
The results demonstrate the superior performance of SIRFP for semantic segmentation tasks. The code is available at \url{https://github.com/dywu98/SIRFP.git}.
\end{abstract}

\section{Introduction}
Semantic segmentation is a classic and challenging task in the field of computer vision. Its goal is to assign category labels to each pixel in an image. Recently, with the advent of Fully Convolutional Networks (FCN)~\cite{fcn}, numerous deep learning-based methods~\cite{chen2017deeplab,ding2018context,unet,psp,pang2019towards,chen2018encoder} have been proposed and applied to semantic segmentation. These methods often involve networks with a large number of parameters and high computational costs, which enable them to extract high-quality features for pixel representation. However, the intimidating computational cost also hinders their deployment on resource-constrained devices widely used on autonomous vehicles, mobile phones, wearable devices, and robots. Therefore, there is an urgent need to develop efficient neural network compression methods for semantic segmentation.

\begin{figure}[t!]
\centering
\begin{center}
\renewcommand{\arraystretch}{0.3} %
\renewcommand{\tabcolsep}{0.3pt} %
\includegraphics[width=0.99\linewidth]{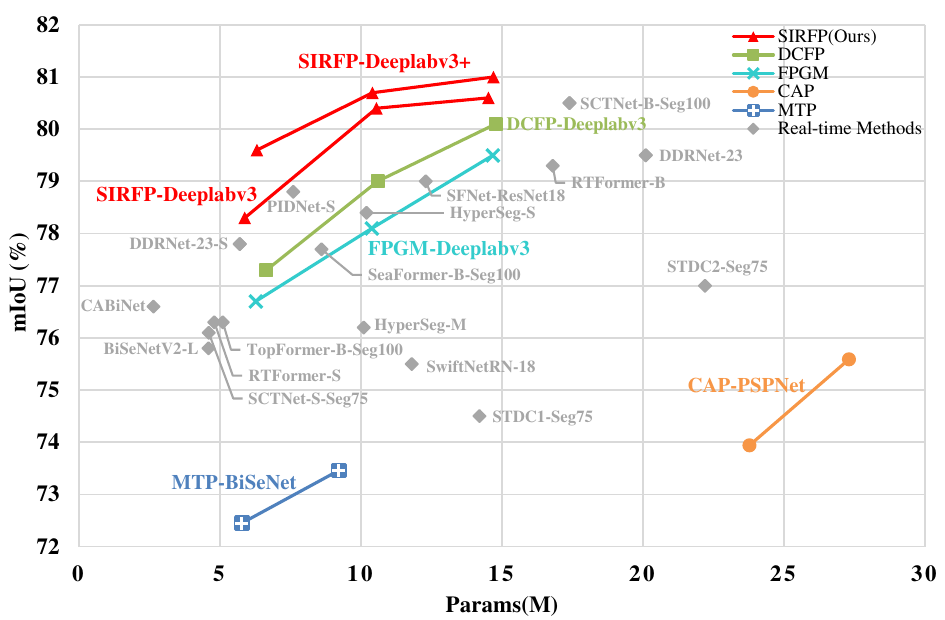} 
\end{center}

\caption{
Params vs mIOU on Cityscapes dataset. The proposed Spatial-aware Information Redundancy Filter Pruning (SIRFP) method achieves an outstanding trade-off between parameter efficiency and accuracy. The pruning methods are connected with lines while real-time methods are colored in grey. Best viewed in color.
}
\label{fig:param}
\end{figure}

Currently, network compression methods can be broadly categorized into three types: parameter quantization~\cite{deep_compression,single_multilevel_quantization,step_size_quantization}, low-rank approximation~\cite{legonet,on_compressing}, and network pruning~\cite{thinet,dynamic_model_pruning,importance_estimation}. Network pruning can be further divided into weight pruning~\cite{learning_weights_connections,Learning-Compression} and filter pruning~\cite{Discrimination-aware,importance_estimation,Pruning_Filters_for_Efficient_ConvNets}. Weight pruning eliminates individual weight elements to achieve unstructured sparsity, requiring specialized libraries to accelerate sparse matrix computations. In contrast, filter pruning directly prunes redundant channels (or filters). The pruned network retains the same structure as the original, differing only in the number of channels. Thus, filter pruning can achieve network compression using standard hardware without specially designed libraries. 
This simplicity has made filter pruning a widely adopted approach, provoking extensive research in this area.

However, most pruning methods are primarily focused on image recognition tasks. 
Although these methods achieve promising performance in image classification tasks, they fail to account for the differences between image recognition and semantic segmentation.
Classification focuses on identifying distinguishing features, whereas segmentation further requires precise object localization within images. Rich spatial information is crucial for high-quality segmentation, but it plays a lesser role in classification. Consequently, semantic segmentation models exhibit different channel redundancy characteristics compared to classification models. Identifying redundancy in segmentation requires evaluating not only the similarity of semantic information but also the spatial redundancy across channels.
The spatial-agnostic importance metric adopted in existing filter pruning methods limits their effectiveness on semantic segmentation models, leading to a suboptimal trade-off between compression ratios and accuracy performance.

To address these issues, we propose a spatially-aware information redundancy filter pruning (SIRFP) method for semantic segmentation.
To measure the importance of each channel, we first introduce a spatially-aware redundancy metric, statistically calculating the redundancy based on the non-pooled, high-resolution, and spatially informative features during training using the exponential moving average (EMA) mechanism. 
With this spatial-aware redundancy metric, we rethink the optimization objective of structured pruning from the graph theory perspective, finding it analogous to the classical \textit{Maximum Edge Weight Clique Problem} (MEWCP)~\cite{mewcp}.
This problem aims to find the complete subgraph with the maximum sum of all its edge weights of a given cardinality~(the number of nodes).
From this perspective, we construct the graph of filters: each node represents a filter, and the edge weights are negatively correlated with the mutual information redundancy between filters.
With this constructed graph, the filter pruning optimization goal that minimizes the mutual information redundancy between channels in the pruned network is exactly equivalent to MEWCP.
Unfortunately, MEWCP is an NP-hard problem, which could not be directly solved in polynomial time.
Although various heuristic methods~\cite{mewcp, pullan2008approximating} and exact methods~\cite{hosseinian2020lagrangian} have been proposed for MEWCP, they still cost unacceptable time in network pruning scenarios where the number of filter nodes is much larger (more than 1024) than the cardinality of graphs (less than 500) that they typically focus on.
To address this challenge, we propose an efficient greedy algorithm that significantly reduces the computational complexity and time cost, providing a practical solution to our revised pruning optimization problem. 
Extensive experiments on popular semantic segmentation datasets and models demonstrate that our method achieves outstanding results.
For example, Figure~\ref{fig:param} shows that the proposed SIRFP achieves the state-of-the-art trade-off between the number of parameters and mIoU performance on the Cityscapes dataset compared with both pruning methods and recent real-time segmentation methods. 
Our method is also evaluated on object detection and image classification tasks, showing decent generalization ability.
In summary, the contributions of this paper are as follows:
\begin{enumerate}
    \item We highlight that current channel pruning methods overlook the spatial redundancy in semantic segmentation models, leading to limited performance and compression ratios.
    \item To address this issue, we propose a spatially-aware redundancy metric that directly calculates information redundancy based on high-resolution feature maps that retain spatial location information, using a sliding average during network training to avoid additional inference and feature storage post-training.
    \item We propose to minimize spatial-aware redundancy between the left channels in the pruned model and frame the optimization objective as the maximum weight clique problem~(MEWCP). A fast and efficient greedy pruning process is presented to solve this NP-hard problem.
\end{enumerate}

\section{Related Work}

\subsection{Semantic Segmentation}

Since FCN~\cite{fcn} introduced a simple and effective solution for semantic segmentation, numerous deep learning methods have continuously pushed the performance boundaries in this field.
Compared with classification methods, segmentation approaches are dedicated to aggregating more contextual information and preserving rich spatial information.
For example, PSPNet~\cite{psp} and the DeepLab series~\cite{chen2017deeplab,chen2017rethinking, chen2014semantic, chen2018encoder} maintain a rather large resolution of feature map for the final prediction.
Recent transformer-based methods like ViT~\cite{vit} and Swin~\cite{liu2021swin} also enhance spatial information by adopting positional encoding to improve the performance.

In the field of real-time segmentation, features with abundant spatial details are even more important given that the computational cost is under constraints.
However, they manage to preserve rich spatial information by adopting light-weighted multi-branch networks.
For instance, the BiSeNet~\cite{yu2018bisenet,yu2021bisenet} series introduces a two-branch architecture to get rich spatial details and high-level semantic information, respectively. 
STDC~\cite{stdc} adopts an additional spatial guidance branch to enhance spatial information in the extracted features.
Recently, DDRNet~\cite{ddrnet}, RTFormer~\cite{wang2022rtformer}, and SeaFormer~\cite{wan2023seaformer} employ a feature-sharing framework which divides spatial and semantic features at the deep stages.
All these methods demonstrate the importance of the spatial distribution information for segmentation.

\subsection{Network Pruning}

Network Pruning is a popular and effective technique for network compression, which aims to eliminate redundant parameters in the neural network.
Opposed to unstructured pruning (weight pruning), structural pruning removes parameters following certain structured sparsity modes.
Filter pruning, which removes the entire filters, is one of the most popular techniques to get an efficient model without the support of specially designed hardware or libraries.
The key is how to evaluate the importance of each filter and the corresponding channel.
Most of them~\cite{geometric, networksliming, depgraph, thinet} resort to the property of filter weights, while there are also methods~\cite{chip, hrank} analyze the characteristic of the corresponding feature maps.
As most of the existing pruning methods focus on image classification, the spatial distribution in the feature map, which is of significance for location-sensitive tasks such as segmentation and detection, hardly attracts the attention of researchers.

Some recent works are designed specifically for semantic segmentation.
CAP~\cite{cap} first proposes to quantify the contextual information using pooled features. The quantified context is utilized as a regularization term.
It penalizes the scaling factors of channels with less contextual information and encourages those with more contextual information.
Chen et al.~\cite{mtp} employ a multi-task channel pruning framework where the importance of each channel is simultaneously determined by both the classification and segmentation tasks.
DCFP~\cite{dcfp} pointed out that the long-tail distribution of samples hinders the performance of pruned networks. Thus, a gradient-based importance criterion along with the powerful Geometric-Semantic Re-balanced Loss (GSRL) is proposed to avoid performance loss and boost the learning on those tail classes during fine-tuning.
These methods significantly improve the efficiency-accuracy trade-off of pruned segmentation models, but still ignore the great disparity between image classification and semantic segmentation: segmentation requires abundant spatial information to locate objects while classification only needs class-related information for categorizing.
Therefore, without rich spatial information, the utilization of only parameters of filters (or channels) or the pooled features prevents them from achieving superb segmentation performance of pruned models.

\section{Methodology}
In this section, we first introduce the optimization problem of most pruning methods. Then, our motivation to address the network pruning issue based on the graph theory is elaborated. After that, we introduce the details of the optimization problem of structural pruning as maximum edge weight clique problem (MEWCP), the proposed spatial-aware information redundancy metric, and our efficient heuristic greedy pruning~(EHGP) method to efficiently address the NP-hard MEWCP.

\subsection{Preliminary}

For clarity, we first define the denotations.
Suppose a neural network contains $L$ layers. The set of network parameters $\{\mathbf{W}^1, \cdots, \mathbf{W}^L \}$ contains all the learnable weights. 
$\mathbf{W}^l \in \mathcal{R}^{C^l \times C^{l-1}}$ denotes the weight matrix of \textit{l}-th convolution or linear layer, where $C^l$ and $C^{l-1}$ denoting the number of output and input channels, respectively. The \textit{i}-th column in $\mathbf{W}^l$, namely the weight vector of the \textit{i}-th output channel, in the $l$-th layer is represented by $\mathbf{w}^l_{i,:}$. 

For the $l$-th layer with channel sparsity $\kappa^l$, channel pruning methods identify and retain the set of channels $\mathcal{T}^l$  containing the most important information to form an efficient light-weighted pruned model. 
The optimization problem can be formulated as follows:
\begin{equation}
\begin{aligned}
\mathop{\max}\limits_{\mathbf{m}^{l}} &\sum_{i=1}^{C^l} m^l_{i}\Phi(l, i), &
\mathit{s.t.} &\sum_{j=1}^{C^l}m^l_{i}=\kappa^l C^l,
\label{eq:pre_prob}
\end{aligned}
\end{equation}
where $\Phi(l, i)$ denotes the importance score of the \textit{i}-th filter in the \textit{l}-th layer, 
$\mathbf{m}^l=[m^l_1, m^l_2, ..., m^l_{C^l}]$ is the mask vector of pruning decision for the l-th layer.
$m^l_{i}$ is set to 0 if the i-th output channel will be pruned or 1 if it will be kept.

\subsection{Motivation}

From the redundancy perspective, the channel~(or filter) set $\mathcal{T}^l$ should contain those channels whose output features are the most informative ones. 
In other words, if the output feature distribution of a channel is similar to those of other channels, the information it contains is redundant. 
In this case, even if this redundant channel is pruned, the next layer can still incorporate similar important information from those channels that survived. 
Intuitively, channels with high redundancy should be removed, while those less redundant ought to be kept since they contain discriminative information which is difficult to reconstruct from other channels.

Some existing methods like FPGM~\cite{geometric} have explored this idea in image recognition tasks.
However, the redundancy~(or channel similarity) is calculated based solely on the weights of channels.
Despite that these methods achieve promising progress on the image classification task, we find that they fail to achieve satisfactory results on the task of location-sensitive tasks such as semantic segmentation and object detection.
This defect can be attributed to the overlook of spatial information.
Evaluating the class-related semantic information using the weight of filters has been proven to be effective in image classification models.
However, these discrepancies can never capture the variations of spatial distribution between different feature maps, which will lead to poorly located objects for semantic segmentation.
The reason for such a performance difference between the pruned classification and segmentation models lies in that classification models typically pool feature maps into vectors, while segmentation models deeply rely on the feature maps with rich spatial information.
Therefore, the pruned classification models can still make correct classification predictions based on these spatially misaligned feature maps.
Nevertheless, for semantic segmentation, the misaligned feature maps directly divert the attention of succeeding layers from the correct position, which eventually results in poor performance.

To address this issue, we propose a spatial-aware channel redundancy pruning method for semantic segmentation.
In contrast to most existing methods that evaluate the importance of channels, some pioneering works such as CHIP~\cite{chip} and HRank~\cite{hrank} propose to compute the importance score in a data-driven manner: the pre-trained network inference on a set of training data to get real feature maps which are then utilized for the calculation of importance score.
Inspired by them, we propose to minimize the feature map information redundancy among the remained channels in the pruned network in a spatial-aware manner.

\subsection{Structural Pruning as Maximum Edge Weight Clique Problem}
From the perspective of redundancy, each remained channel should produce exclusive discriminate information to maintain low redundancy in the pruned model.
In other words, all $\kappa^l C^l$ channels are selected and kept from all the $C^l$ ones such that their information redundancy (or similarity) between them is minimized.
With this goal, the channel pruning problem can be naturally represented as the \textit{Maximum Edge Weight Clique Problem} (MEWCP) with a cardinality bound.
This provides us with a new guideline to design pruning methods in a principled way based on the overall perspective of all the remained channels and graph theory knowledge.

\noindent \textbf{Definition 1} (\textit{Maximum Edge Weight Clique Problem}). Let $G=(V, E)$ be a simple, complete, undirected, and edge-weighted graph, where $V=\{1,2,...,C^{l}\}$ is the set of channel vertices (nodes) and $E$ is the set of edges between channel vertices. The clique is defined as a subset of vertices $C\subseteq V$ which is an induced subgraph of $G$ that is complete, i.e. for all vertices $i,j \in C$, edge $\{i, j\}$ exist in the corresponding subgraph $G[C]$.
The weight of the edge $\{i, j\}$ is denoted as $a_{ij}$, which is negatively correlated with the proposed spatial-aware information redundancy metric.
The summation of edge weights of clique $C$ is noted as $W^{edge}_C$.
The maximum edge weight clique of G is a clique $C$ in $G$ which maximizes $W^{edge}_C$.
The MEWCP with a cardinality bound constraint aims to find the maximum edge weight clique with no larger than $b$ vertices, which be formulated as the follows:
\begin{equation}
\begin{aligned}
\mathop{\max}\limits_{\mathbf{m}^l} &\sum_{i\in V}\sum_{j\in V \backslash \{i\}}  m^l_{i} m^l_{j} a_{ij}, &
\mathit{s.t.} &\sum_{i\in V} m^l_{i} = b.
\end{aligned}
\end{equation}
When $b=\kappa^l C^l$, taking channels as vertices, and set $\Phi(l, i)=\sum_{j\in V \backslash \{i\}} m^l_{j} a_{ij}$, this problem has the same format with that of channel pruning. The details of edge weight $a_{ij}$ are introduced in the next subsection.

\subsection{Spatial-aware Information Redundancy Metric}

To evaluate the information redundancy and keep the spatial information, we utilize the un-pooled feature map for calculation.
We have to point out that previous data-driven pruning methods, like HRank and CHIP, only use the pooled vector for importance score calculation, which completely eliminates the spatial information.
For semantic segmentation, spatial information in the feature map is vital, but not taken into consideration by existing pruning methods.
Thus, we keep each element in the feature map, preserving the abundant location information for redundancy metric to enable the spatial-aware ability.
Here, we propose the redundancy metrics which can be formulated as:
\begin{equation}
\begin{aligned}
r_{ij}=log2-\frac{1}{2}D_{KL}(\mathbf{F}^l_i||\frac{\mathbf{F}^l_i+\mathbf{F}^l_j}{2}) - \frac{1}{2}D_{KL}(\mathbf{F}^l_j||\frac{\mathbf{F}^l_i+\mathbf{F}^l_j}{2}),
\label{eq:js}
\end{aligned}
\end{equation}
where 
$D_{KL}(\cdot)$ denotes the Kullback-Leibler divergence,
$\mathbf{F}^l_i, \mathbf{F}^l_j \in \mathcal{R}^{C^l \times H \times W}$ denote the un-pooled output feature map of the i-th and j-th filter with spatial resolution $H\times W$, respectively.

As our optimization goal is to maximize the summation of edge weights, the edge weight should be negatively correlated with the redundancy.
Moreover, if the calculation is conducted after pre-training, it would cost a huge memory and time to inference on real training data and collect the feature maps.
Thus, we adopt the Exponential Moving Average (EMA) mechanism to accumulate the edge weight since the calculation of $r_{ij}$ requires high-resolution feature maps of each layer.
Finally, the edge weight $a_{ij}$ can be defined as the follows:
\begin{equation}
\begin{aligned}
a^t_{ij} = a^{t-1}_{ij} \alpha + (1-r_{ij})(1-\alpha),
\label{eq:edge_weight}
\end{aligned}
\end{equation}
where $\alpha$ is the hyper-parameter of EMA, which is set to 0.99.
Hence, with this redundancy metric and edge weight, we can leverage the rich theoretical foundations of MEWCP to derive a spatial-aware pruning method.

\subsection{Efficient Heuristic Greedy Pruning}
Unfortunately, an practical issue arises when it comes to address this problem in such a channel pruning scenario: MEWCP is apparently NP-hard~\cite{hosseinian2017maximum}, which requires a huge amount of time to get the exact solution.
The large number of channels and layers makes it even more time-consuming, since we have to consider all the  $b\choose {C^l} $ possibilities for each layer. 
Although there researchers attempt to reduce the computational complexity, acquiring an exact solution is still too expensive compared to existing channel pruning methods whose complexity is usually only $\mathcal{O}(C^l+C^l log(C^l))$.
In addition, the heuristic methods for MEWCP is also not robust enough.
Studies shows that the computational complexity of some heuristic methods~\cite{macambira2000edge} could reach $\mathcal{O}(n^4)$ or even $\mathcal{O}(n^5)$~\cite{hosseinian2017maximum}.
Therefore, it is imperative to design a computational efficient algorithm for our MEWCP pruning method.

Although the cost for exactly solving MEWCP is unacceptable for most case, the solution for $b=C^l-1$ can be explicitly given~\cite{mewcp}.
Putting $\{k\}:=V\backslash C$, the edge weight of clique $C$ can be formulated as :
\begin{equation}
\begin{aligned}
W^{edge}_C&=\sum_{i\in C}\sum_{j\in C \backslash \{i\}}a_{ij} = 
\sum_{i\in V}\sum_{j\in V \backslash \{i\}}  m^l_{i} m^l_{j} a_{ij} \\
&=\sum_{i\in V}\sum_{j\in V \backslash \{i\}}a_{ij} - \sum_{h \in V\backslash \{k\}}a_{hk}.
\end{aligned}
\end{equation}
Thus, $C$ can be found by only excluding $k$ out of $V$ such that the vertices $k$ has the minimum summation of edge weights in $V$:
\begin{equation}
\begin{aligned}
\mathop{\min}\limits_{k} \sum_{h \in V\backslash \{k\}}a_{hk}.
\end{aligned}
\end{equation}
Hence, if we prune only one channel out of the l-th layer, the original optimization problem is equivalent to selecting the node $k$ with the minimum sum of all its edge weights.

\begin{algorithm}[t!] 
  \caption{Efficient Heuristic Greedy Pruning}
  \label{alg:ours}
  \SetKwInput{KwIn}{Input~~~}
  \SetKwComment{Comment}{}{}
  \DontPrintSemicolon
  
  \KwIn{Weight of the l-th layer $\mathbf{W}^l \in \mathcal{R}^{C^{l} \times C^{l-1}} $ where $\mathbf{w}^l_{i,:}$ is the weight of $i$-th channel, 
  the graph $G=(V,E)$, 
  and the channel vertices set $V=\{ 1,2,...,C^l\}$}

  \KwOut{The weight $\mathbf{W}^l_p \in \mathcal{R}^{C^{l}_p \times C^{l-1}}$ after pruning}
  \smallskip
  Get the edge weight matrix $\mathbf{A}^l$ \\
  Set the vertices set $V_r$ to be removed as empty set \\
  Set the pruning index $k=1$, counter $t=1$ \;
  \Comment*[l]{\textcolor{lightgray}{\small // Calculate sum of edge weight:}} 
  \For{$i \in V$} 
  {
    $s_i \leftarrow \sum_{j \in V\backslash \{i\}}a_{ij}$ \\
    \If{$s_{k} \textgreater s_i$}
    {
        $k \leftarrow i$ \Comment*{\textcolor{lightgray}{\small // Get the least sum}}
    }
  } 
  \Comment*[l]{\textcolor{lightgray}{\small // Start iterative pruning:}} 
  \For{each prune iteration $t_p \in [b] $}
  {
      $V_r \leftarrow V_r+\{k\}$ \Comment*{\textcolor{lightgray}{\small // Prune the selected k}}
      $q \leftarrow random(V-V_r)$ \\
      \For{$i \in V-V_r$} 
      {
        \Comment*[l]{\textcolor{lightgray}{\small // Update sum for next iteration}}
        $s_i \leftarrow s_i-a_{ik}$ \\
        \If{$s_{q} \textgreater s_i$}
        {
            $q \leftarrow i$ \Comment*{\textcolor{lightgray}{\small // Get the least sum}}
        }
      } 
      $k\leftarrow q$;~~$t\leftarrow t+1$
  }
  $\mathbf{W}^l_p \leftarrow$ $Concate($ [$\mathbf{w}^l_{i:}$, \textbf{for} $i\in V-V_r$] $)$ \\
  
  \Return $\mathbf{W}^l_p$
\end{algorithm}

Therefore, we propose a greedy algorithm called efficient heuristic greedy pruning (EHGP) to reduce the complexity of solving MEWCP, as shown in Algorithm 1. 
We decompose the original problem of pruning $C^l-b$ channels(nodes) at once into the task of iteratively deleting a single channel for $C^l-b$ times.
First, we calculate the summation of edge weights for each vertices.
Then, during each iteration in the EHGP, we prune the single channel vertices with the least sum of edge weights for all layers and update the sum of left edge weights of those still remaining vertices for the next iteration.
In this greedy way, our proposed method only needs $b$ iterations to solve the MEWCP in such a network pruning scenario. 
Thus, the average complexity of the proposed pruning procedure is reduced to $\mathcal{O}(n^2)$.
We also compare the runtime of our method and other MEWCP solutions to prune a Deeplabv3-ResNet50 model. The results in Table~\ref{tab:runtime} demonstrate the efficiency of EHGP.

\subsection{Pruning Pipeline}
As we utilize the feature maps to acquire the information redundancy, a simple way to achieve this is to infer on a set of selected data after the pre-training stage to collect feature maps, which will lead to undesired additional computation costs.
Some recent works~\cite{hrank,chip} adopt this \textit{pretrain-collect-prune-finetune} pipeline.
Accordingly, they suffer from extremely large time costs (more than an hour on ImageNet~\cite{imagenet}) to perform the \textit{collect\&prune}.
Inspire by previous works~\cite{bn}
we adopt the Exponential Moving Average (EMA) mechanism to accumulate the edge weight matrix $\mathbf{A}^{l}$ during training, as shown in Equation~\ref{eq:edge_weight}.
In order to mitigate the large performance drop after pruning, we progressively remove channels in $T_{step}$ iterations.
We prune the model and finetune it for a few iterations to restore the performance, which makes the pruning less destructive.
This can also reduce the duration of accumulating edge weight matrix and lead to an accurate reflection of the feature redundancy between channels.
The detailed overall pruning process is shown in the Algorithm~\ref{alg:overall}.

\begin{algorithm}[t!] %
  \caption{Overall Pruning Pipeline}
  \label{alg:overall}
  \SetKwInput{KwIn}{Input~~~}
  \SetKwComment{Comment}{}{}
  \DontPrintSemicolon
  
  \KwIn{An $L$-layer model with all the weights $\Theta=\{\mathbf{W}^1, \cdots, \mathbf{W}^L \}$, 
  and the training dataset $D$ }

  \KwOut{The pruned and finetuned model $\Theta'_p$}

  \smallskip
  \smallskip

  \For{each pre-training iteration $t \in [T^{pre}_{total}]$}
  {
    Sample a mini-batch ($\mathbf{X}$,$\mathbf{Y}$) from $D$ \;
    Calculate spatial-aware redundancy based on feature maps \Comment*[r]{\small \textcolor{lightgray}{// Equation 3}} 
    
    Update edge weight matrix $\mathbf{A}^l_t$ of each layer using EMA \Comment*[r]{\small \textcolor{lightgray}{// Equation 4}}
    Update the unpruned model $\Theta$  \;
  }

  $\Theta^{prev} \leftarrow \Theta$ \;
  
  \For{each progressive pruning step $t_s \in [T_{step}]$}
   {
      Prune $\Theta^{prev}$ by EHGP in Algorithm~\ref{alg:ours} and get pruned model $\Theta^{t_s}_p=\{\mathbf{W}^1_p, \cdots, \mathbf{W}^L_p \}^{t_s}$ \\
    
      \For{each fine-tuning iteration $t \in [T^{fine}_{total}]$}
      {
        Sample a mini-batch ($\mathbf{X}$,$\mathbf{Y}$) from $D$ \;
        Update the weight of pruned model $\Theta^{t_s}_p$ \;
      }
      Set the pruned and finetuned model $\Theta^{t_s}_f$ as the initial model for the next step: $\Theta^{prev}=\Theta^{t_s}_f$ \;
    }
  \Return the final pruned and finetuned model $\Theta^{T_{step}}_f$.
\end{algorithm}

\section{Experiments and Analysis}
Although this paper mainly focuses on semantic segmentation, to comprehensively evaluate the efficacy and generality of our method, we also conduct experiments on a variety of computer vision tasks, including image classification and object detection.
The experiments run on the Pytorch framework using NVIDIA RTX 3090 GPUs. 
More implementation details, ablation studies, and visualization results are presented in the Appendix.

\subsection{Datasets}
In this work, we conduct experiments on three tasks: semantic segmentation, image classification, and object detection.
Following DCFP, aside from the popular but rather easy Cityscapes~\cite{cordts2016cityscapes} dataset (19 categories), we also evaluate our method on the challenging ADE20k~\cite{ade} (150 categories) and COCO stuff-10K~\cite{coco} (171 categories).
For image classification and object detection, we compare the proposed method with other state-of-the-art pruning methods on ImageNet~\cite{imagenet} and COCO2017 dataset~\cite{coco2017}, respectively.

\begin{table}[t!]

  \caption{Comparison with other pruning methods on Cityscapes validation set using Deeplabv3-ResNet50.}
  \label{tab:city}
  \centering
  \small
  \resizebox{0.99\linewidth}{!}{
  \begin{tabular}{ l c c c c c c c c c}
      \toprule
      \bf Method & \bf mIoU(\%) & \bf Params(\%$\downarrow$) & \bf FLOPs(\%$\downarrow$) \\
      \midrule
      Unpruned   & 81.6 & 41.27M & 1418.29G \\
      \midrule
      Random     & 78.7 & 16.60M~(59.78\%$\downarrow$) & 558.72G~(60.61\%$\downarrow$) \\
      NS         & 79.9 & 15.57M~(62.27\%$\downarrow$) & 575.20G~(59.44\%$\downarrow$) \\
      Taylor     & 80.3 & 14.99M~(63.68\%$\downarrow$) & 564.88G~(60.17\%$\downarrow$) \\
      DepGraph   & 80.0 & 16.84M~(59.20\%$\downarrow$) & 561.92G~(60.38\%$\downarrow$) \\
      FPGM-60\%   & 80.2 & 14.68M~(64.43\%$\downarrow$) & 550.75G~(61.17\%$\downarrow$) \\
      DCFP-60\%   & 80.9 & 14.78M~(64.19\%$\downarrow$) & 553.74G~(60.96\%$\downarrow$) \\
      \bf Ours-60\%   &\bf 81.3 & \bf 14.52M~(64.82\%$\downarrow$) & \bf  549.84G~(61.23\%$\downarrow$) \\
      \midrule
      FPGM-70\%   & 79.3 & 10.51M~(74.82\%$\downarrow$) & 411.26G~(71.00\%$\downarrow$) \\
      DCFP-70\%   & 79.8 & 10.60M~(74.32\%$\downarrow$) & 418.08G~(70.52\%$\downarrow$) \\
      \bf Ours-70\%   &\bf 80.9 &\bf 10.45M~(74.44\%$\downarrow$) & \bf  398.56G~(71.90\%$\downarrow$)\\
      \midrule
      FPGM-80\%   & 77.9 & 6.27M~(84.81\%$\downarrow$) & 274.18G~(80.67\%$\downarrow$) \\
      DCFP-80\%   & 78.8 & 6.65M~(83.89\%$\downarrow$) & 280.96G~(80.19\%$\downarrow$) \\
      \bf Ours-80\%   &\bf 79.4 &\bf 5.88M~(85.75\%$\downarrow$) & \bf 260.99G~(81.60\%$\downarrow$) \\
      \bottomrule
  \end{tabular} }

\end{table} 

\begin{table}[t!]

  \caption{Ablation studies on Deeplabv3+ and PSPNet. All models are trained on Cityscapes. }
  \label{tab:networks}
  \centering
  \small
  \resizebox{0.99\linewidth}{!}{
  \begin{tabular}{ l c c c c }
      \toprule
       \multirow{2}{*}{\bf Method} & \multicolumn{2}{c}{\bf Deeplabv3+} & \multicolumn{2}{c}{\bf PSPNet}\\
       \cmidrule{2-5}
        & \bf mIoU(\%) & \bf FLOPs(\%$\downarrow$) & \bf mIoU(\%) & \bf FLOPs(\%$\downarrow$) \\
      \midrule
      Unpruned  & 81.9 & 1608.99G & 81.3 & 1476.54G \\
      \midrule
      Random    & 76.4  & 666.40G & 76.8 & 582.08G \\
      DCFP-60\%  & 80.6 & 652.08G & 79.1 & 595.04G \\
       \bf Ours-60\%  & \bf 81.6 & \bf 614.81G & \bf 81.1 & \bf 573.96G \\
      \midrule
      DCFP-70\%  & 79.5 & 502.32G & 78.1 & 444.16G \\
      \bf Ours-70\%  & \bf 81.2 & \bf 460.02G & \bf 80.2 & \bf 412.52G \\
      \midrule
      DCFP-80\%  & 78.4 &  330.96G  & 76.0 & 314.40G \\
      \bf Ours-80\%  &\bf 80.3  &\bf 306.01G &\bf 79.6 & \bf 286.93G \\
      \bottomrule
  \end{tabular} }
\end{table}

\subsection{Semantic Segmentation}
\noindent \textbf{Comparison on Cityscapes} To compare our method with other popular pruning algorithms including NS~\cite{networksliming}, Taylor~\cite{importance_estimation}, Depgraph~\cite{depgraph}, FPGM~\cite{geometric}, and DCFP~\cite{dcfp}, we reproduce these methods and prune the DeeplabV3-resnet50 model on Cityscapes. 
The results are shown in Table~\ref{tab:city}.
Following DCFP, all our experiments adopt the GSRL to improve performance.
Our method attains a significant performance advantage over the other pruning methods.
Moreover, we also include recent real-time segmentation methods and more pruning methods for segmentation like CAP~\cite{cap} and MTP~\cite{mtp} in Figure~\ref{fig:param} to compare the number of parameters and mIoU performance.
According to the results, SIRFP achieves the best parameter-mIoU trade-off among all the other pruning methods and real-time methods.

\begin{table}[t!]
  \caption{Results comparison with other pruning methods of Deeplabv3-ResNet50 on ADE20K and COCO stuff-10K.}
  \label{tab:dataset}
  \centering
  \small
  \resizebox{0.96\linewidth}{!}{
  \begin{tabular}{ l c c c c }
      \toprule
       \multirow{2}{*}{\bf Method} & \multicolumn{2}{c}{\bf ADE20K} & \multicolumn{2}{c}{\bf COCO stuff-10k}\\
       \cmidrule{2-5}
        & \bf mIoU(\%) & \bf FLOPs & \bf mIoU(\%) & \bf FLOPs \\
      \midrule
      Unpruned  & 45.3 & 177.42G & 38.1 & 177.45G \\
      \midrule
      Random    & 40.2 & 71.29G & 28.3 & 70.59G \\
      NS        & 40.6 & 72.33G & 31.1 &  71.61G \\
      Taylor    & 40.8 & 73.83G & 32.0 & 70.89G \\
      DepGraph  & 41.0 & 70.31G & 32.0 & 70.89G \\
      FPGM  & 39.8 & 72.44G & 30.3 & 72.44G \\
      DCFP-60\%  & 44.8 & 71.10G & 34.3 & 70.45G \\
      \bf Ours-60\%  & \bf 44.9 & \bf{67.82}G & \bf 34.8 & \bf 69.65G \\
      \midrule
      DCFP-70\%  & 44.4 & 55.93G & 33.1 & 51.92G \\
      \bf Ours-70\%  & \bf 44.6 & \bf{51.40}G & \bf 33.9 & \bf 50.74G \\
      \midrule
      DCFP-80\%  & 41.4 & 35.96G & 30.3 & 35.83G \\
      \bf Ours-80\%  & \bf 41.8 & \bf{32.39}G & \bf 30.7 & \bf 33.25G \\
      \bottomrule
  \end{tabular} }
\end{table} 

\begin{table}[t!]
  \caption{Object detection pruning results on COCO2017 validation set using SSD-ResNet50. * denotes our reproduction.}
  \label{tab:detection}
  \centering
  \small
  \resizebox{0.98\linewidth}{!}{
  \begin{tabular}{l  c c c c c c c}
      \toprule
      \multirow{2}{*}{\bf Method} & \bf FLOPs  & \multirow{2}{*}{$\mathbf{AP}$} & \multirow{2}{*}{ $\mathbf{AP}_{50}$} & \multirow{2}{*}{$\mathbf{AP}_{75}$} & \multirow{2}{*}{$\mathbf{AP}_{S}$} & \multirow{2}{*}{$\mathbf{AP}_{M}$} & \multirow{2}{*}{$\mathbf{AP}_{L}$} \\
      & \bf Reduction & \\
      \midrule
      Baseline & 0\% & 25.2 & 42.7 & 25.8 & 7.3 & 27.1 & 40.8 \\
      \midrule
      DMCP     & 50\% & 24.1 & 41.2 & 24.7 & 6.7 & 25.6 & 39.2 \\
      FPGM\textsuperscript{*}   & 50\% & 25.2 & 42.4 & 25.9 & 7.0 & 27.2 & 41.3 \\
      CHEX-1   & 50\% & 25.9 & 43.0 & 26.8 & 7.8 & 27.8 & 41.7 \\
      \bf Ours & 50\% & \bf 26.6 & \bf 44.2 & \bf 27.3 & \bf 8.1 & \bf 28.8 & \bf 43.4 \\
      \midrule
      CHEX-2   & 75\% & 24.3 & 41.0 & 24.9 & 7.1 & 25.6 & 40.1 \\
      \bf Ours & 75\% & \bf 24.6 & \bf 41.8 & \bf 25.1 & \bf 7.3 & \bf 26.0 & \bf 41.2 \\
      \bottomrule
  \end{tabular} 
  }
\end{table} 

\noindent \textbf{Comparison on ADE20K and COCO stuff-10k} The comparison with existing pruning methods on ADE20K and COCO stuff-10K are reported in Table~\ref{tab:dataset}, showing consistent effectiveness on these challenging datasets. 
Specifically, our method exceeds the FPGM by 1.1\%, 5.0\%, and 4.5\%  mIoU under 60\% FLOPs reduction on Cityscapes, ADE20K, and COCO stuff-10k, respectively.
We attribute this significant performance advantage to the proposed spatial-aware redundancy metric, which enables our method to capture the overlap of spatial distribution between different features.

\noindent \textbf{Comparison with DCFP} As DCFP~\cite{dcfp} also calculates the importance score using the EMA mechanism, the performance advantage of our method over DCFP indicates that the adoption of the EMA mechanism is obviously not the main reason for performance advantage. 
We believe that this corroborates our argument that the spatial-aware metric is more suitable for semantic segmentation than traditional spatial-agnostic metrics, like the grad-based importance score of DCFP.

\subsection{Object Detection}
Object detection is also a location-sensitive computer vision task.
We also conduct experiments on the classic SSD-ResNet50 and compare the results with DMCP~\cite{guo2020dmcp}, FPGM~\cite{geometric}, CHEX~\cite{chex} to evaluate the effectiveness and generalization of our method.
Experimental results conducted on the COCO2017 dataset are reported in Table~\ref{tab:detection}.
Following CHEX, we only prune the backbone of SSD.
Compared with the data-driven method CHEX which relies on the rank of the feature map, our method achieves a 0.7\% AP improvement over CHEX under 50\% FLOPs reduction ratio.
The performance advantages over our reproduced FPGM and CHEX illustrate that our spatial-aware method is also more powerful than the other spatial-agnostic pruning methods on object detection task which also requires excellent localization ability.

\subsection{Image Classification}
Although we focus on location-sensitive tasks, the performance on image classification also indicates the ability to preserve spatial-agnostic semantic information.
Hence, we conduct experiments on the ImageNet dataset and the results are reported in Table~\ref{tab:imagenet}.
Following the adaptive pruning methods, we prune the model during training progressively.
SIRFP obtains a significant performance advantage over existing pruning methods.
As CHIP~\cite{chip} and HRank~\cite{hrank} also utilize features to calculate importance score, we contend that the spatial-aware redundancy metric, instead of the data-driven manner, is the main cause of our performance advantage.

\begin{table}[t!]
  \caption{Comparison with existing pruning methods on ImageNet validation set using ResNet-50.}
  \label{tab:imagenet}
  \centering
  \small
    \resizebox{\linewidth}{!}{
    \begin{tabular}{l|cc|c}
    \toprule 
    \textbf{Method} & \textbf{Top-1} & \textbf{Top-5} & \textbf{FLOPs} \\
    \midrule
      SSS~\cite{sss}  & 74.18 & 91.91 & 2.8G \\
      GReg-2~\cite{greg}  & 75.36 & - & 2.8G \\
      AOFP-C1~\cite{aofp}  & 75.63 & 92.69 & 2.6G \\
      ThiNet~\cite{thinet}  & 72.04 & 90.67 & 2.4G  \\
      SFP~\cite{sfp}  & 74.61 & 92.87 & 2.4G \\
      FPGM~\cite{geometric}  & 75.50 & 92.63 & 2.4G \\
      BNFI~\cite{bnfi}  & 75.47 & - & 2.3G \\
      TAS~\cite{tas}  & 76.20 & - & 2.3G \\
      HRank~\cite{hrank}  & 74.98 & 92.33  & 2.3G \\
      Taylor~\cite{importance_estimation} & 74.50 & - & 2.3G \\
      CCP~\cite{ccp}  & 75.50 & 92.62 & 2.1G \\
      GFP~\cite{gfp}  & 76.42 & - & 2.0G \\
      DepGraph~\cite{depgraph}  & 75.28 & - & 2.0G \\
      DSA~\cite{dsa} & 74.69 & 92.06 & 2.0G \\
      CafeNet~\cite{cafenet} & 76.90 & 93.30 & 2.0G \\
      \bf Ours & \bf 77.35 & \bf 93.60 & \bf 2.0G \\
    \cmidrule{1-4}
      BNFI~\cite{bnfi}  & 75.02 & - & 1.9G \\
      FPGM~\cite{geometric}  & 74.83 & 92.32 & 1.9G \\
      Hinge~\cite{hinge}  & 74.70 & - & 1.9G \\
      DCP~\cite{Discrimination-aware}  & 74.95 & \bf 93.54 & 1.8G \\
     \bf Ours & \bf 75.14 & 93.12 & \bf 1.7G \\
     \cmidrule{1-4}
      HRank~\cite{hrank} & 71.98 & 91.01 & 1.6G \\
      DMCP~\cite{guo2020dmcp} & 74.10 & - & 1.1G \\ 
      CHIP~\cite{chip} & 73.30 & 91.48 & 1.0G \\
      \bf Ours & \bf 74.67 & \bf 92.19 & \bf 1.0G\\
    \bottomrule
    \end{tabular}}

\end{table} 

\subsection{Ablation Studies}

\textbf{Generalization on different models.} To evaluate the effectiveness and generalization ability of our pruning method, we conduct experiments on various segmentation models.
Aside from Deeplabv3~\cite{chen2017deeplab}, we also conduct experiments on the popular and classic PSPNet~\cite{psp} and Deeplabv3+~\cite{chen2018encoder}, as reported in the Table~\ref{tab:networks}.
The experimental results demonstrate the decent generalization ability of SIRFP on various models.

\noindent \textbf{Ablation on MEWCP solver.} In order to show the efficiency of our greedy pruning algorithm, we compare the runtime with other exact and heuristic methods for MEWCP in the pruning scenario.
The results in Table~\ref{tab:runtime} show that our greedy algorithm exhibits outstanding efficiency compared with other solutions.
With only $\mathcal{O}(n^2)$ complexity as we discussed above, the speed of the proposed EHGP significantly exceeds other methods for MEWCP.

\begin{table}[t!]
  \caption{Runtime comparison between our proposed EHGP and different methods for the MEWCP. Experiments are conducted on Deeplabv3-ResNet50.}
  \label{tab:runtime}
  \centering
  \small
    \resizebox{\linewidth}{!}{
    \begin{tabular}{lccc}
    \toprule
     \textbf{Method} & \textbf{Type} & \textbf{Runtime(sec.)} & mIoU \\
    \midrule  
      \cite{pullan2008approximating}  & Heuristic & 3731 & 81.3 \\
     \cite{macambira2000edge} & Heuristic & 4933 & 81.1 \\
     \cite{hosseinian2020lagrangian}  & Exact     & \textgreater 10800 & 81.4 \\
     \bf EHGP(Ours)  & Heuristic & 285 & 81.3  \\
    \bottomrule
    \end{tabular}}

\end{table} 

\noindent \textbf{Time efficiency comparison}
We compare the inference time efficiency of DCFP and our SIRFP in Table~\ref{tab:fps}.
All our models employ TensorRT, the commonly used SDK for real-time neural network inference. 
Compared with DCFP, our method performs more efficiently with higher mIoU results and faster inference speed, especially on larger FLOPs reduction ratios. 

\begin{table}[t!]

  \caption{Performance and inference speed comparison with DCFP on Cityscapes dataset using Deeplabv3-ResNet50. }
  \label{tab:fps}
  \centering
  \small
    \begin{tabular}{lccc}
    \toprule
     \textbf{Method}  & \bf \makecell{FLOPs  Reduction} & \textbf{FPS}(img/s) & \textbf{mIoU}(\%) \\
    \midrule  
     Unpruned &  0\%  & 36.7 & 79.3 \\
     \midrule
      DCFP & 60\% & 61.7 & 80.7 \\
    \bf Ours & 60\% & \bf 62.4 & \bf 81.3 \\
    \midrule
      DCFP & 70\%  & 69.9 & 79.8 \\
      \bf Ours & 70\% & \bf 73.9 & \bf 80.9 \\
      \midrule
      DCFP & 80\%  & 81.1 & 78.8 \\
      \bf Ours & 80\% & \bf 90.5 & \bf 79.4 \\

    \bottomrule
    \end{tabular}

\end{table} 

\section{Limitation}
Although SIRFP exhibits excellent performance and generalization ability, it relies on the EMA to accumulate edge weight matrix during training.
For a given well-trained network, the SIRFP has to inference on training samples to acquire feature maps, causing extra computational cost.
Moreover, the time to calculate redundancy may be huge for networks whose feature map resolution is extremely large. 
Thus, the efficiency of redundancy calculation can be further improved.

\section{Acknowledgments}
This work was supported by the Hubei Provincial Natural Science Foundation of China No.2022CFA055, the National Natural Science Foundation of China No.62176097, and  Interdisciplinary Research Program of HUST No.2024JCYJ034.

\bibliography{aaai25}

\begin{thebibliography}{61}
\providecommand{\natexlab}[1]{#1}

\bibitem[{Caesar, Uijlings, and Ferrari(2018)}]{coco}
Caesar, H.; Uijlings, J.; and Ferrari, V. 2018.
\newblock COCO-Stuff: Thing and Stuff Classes in Context.
\newblock In \emph{CVPR}, 1209--1218.

\bibitem[{Carreira-Perpin{\'a}n and Idelbayev(2018)}]{Learning-Compression}
Carreira-Perpin{\'a}n, M.~A.; and Idelbayev, Y. 2018.
\newblock “learning-compression” algorithms for neural net pruning.
\newblock In \emph{CVPR}, 8532--8541.

\bibitem[{Chen et~al.(2014)Chen, Papandreou, Kokkinos, Murphy, and Yuille}]{chen2014semantic}
Chen, L.-C.; Papandreou, G.; Kokkinos, I.; Murphy, K.; and Yuille, A.~L. 2014.
\newblock Semantic image segmentation with deep convolutional nets and fully connected crfs.
\newblock \emph{arXiv preprint arXiv:1412.7062}.

\bibitem[{Chen et~al.(2017{\natexlab{a}})Chen, Papandreou, Kokkinos, Murphy, and Yuille}]{chen2017deeplab}
Chen, L.-C.; Papandreou, G.; Kokkinos, I.; Murphy, K.; and Yuille, A.~L. 2017{\natexlab{a}}.
\newblock Deeplab: Semantic image segmentation with deep convolutional nets, atrous convolution, and fully connected crfs.
\newblock \emph{TPAMI}, 40(4): 834--848.

\bibitem[{Chen et~al.(2017{\natexlab{b}})Chen, Papandreou, Schroff, and Adam}]{chen2017rethinking}
Chen, L.-C.; Papandreou, G.; Schroff, F.; and Adam, H. 2017{\natexlab{b}}.
\newblock Rethinking atrous convolution for semantic image segmentation.
\newblock \emph{arXiv preprint arXiv:1706.05587}.

\bibitem[{Chen et~al.(2018)Chen, Zhu, Papandreou, Schroff, and Adam}]{chen2018encoder}
Chen, L.-C.; Zhu, Y.; Papandreou, G.; Schroff, F.; and Adam, H. 2018.
\newblock Encoder-decoder with atrous separable convolution for semantic image segmentation.
\newblock In \emph{ECCV}, 801--818.

\bibitem[{Chen, Zhang, and Wang(2022)}]{mtp}
Chen, X.; Zhang, Y.; and Wang, Y. 2022.
\newblock MTP: multi-task pruning for efficient semantic segmentation networks.
\newblock In \emph{ICME}, 1--6.

\bibitem[{Cordts et~al.(2016)Cordts, Omran, Ramos, Rehfeld, Enzweiler, Benenson, Franke, Roth, and Schiele}]{cordts2016cityscapes}
Cordts, M.; Omran, M.; Ramos, S.; Rehfeld, T.; Enzweiler, M.; Benenson, R.; Franke, U.; Roth, S.; and Schiele, B. 2016.
\newblock The cityscapes dataset for semantic urban scene understanding.
\newblock In \emph{CVPR}, 3213--3223.

\bibitem[{Deng et~al.(2009)Deng, Dong, Socher, Li, Li, and Fei-Fei}]{imagenet}
Deng, J.; Dong, W.; Socher, R.; Li, L.-J.; Li, K.; and Fei-Fei, L. 2009.
\newblock Imagenet: A large-scale hierarchical image database.
\newblock In \emph{CVPR}, 248--255.

\bibitem[{Ding et~al.(2018)Ding, Jiang, Shuai, Liu, and Wang}]{ding2018context}
Ding, H.; Jiang, X.; Shuai, B.; Liu, A.~Q.; and Wang, G. 2018.
\newblock Context contrasted feature and gated multi-scale aggregation for scene segmentation.
\newblock In \emph{CVPR}, 2393--2402.

\bibitem[{Ding et~al.(2019)Ding, Ding, Guo, Han, and Yan}]{aofp}
Ding, X.; Ding, G.; Guo, Y.; Han, J.; and Yan, C. 2019.
\newblock Approximated oracle filter pruning for destructive cnn width optimization.
\newblock In \emph{ICML}, 1607--1616.

\bibitem[{Dong and Yang(2019)}]{tas}
Dong, X.; and Yang, Y. 2019.
\newblock Network pruning via transformable architecture search.
\newblock \emph{NeurIPS}, 32.

\bibitem[{Dosovitskiy et~al.(2020)Dosovitskiy, Beyer, Kolesnikov, Weissenborn, Zhai, Unterthiner, Dehghani, Minderer, Heigold, Gelly et~al.}]{vit}
Dosovitskiy, A.; Beyer, L.; Kolesnikov, A.; Weissenborn, D.; Zhai, X.; Unterthiner, T.; Dehghani, M.; Minderer, M.; Heigold, G.; Gelly, S.; et~al. 2020.
\newblock An image is worth 16x16 words: Transformers for image recognition at scale.
\newblock \emph{arXiv preprint arXiv:2010.11929}.

\bibitem[{Esser et~al.(2019)Esser, McKinstry, Bablani, Appuswamy, and Modha}]{step_size_quantization}
Esser, S.~K.; McKinstry, J.~L.; Bablani, D.; Appuswamy, R.; and Modha, D.~S. 2019.
\newblock Learned step size quantization.
\newblock \emph{arXiv preprint arXiv:1902.08153}.

\bibitem[{Fan et~al.(2021)Fan, Lai, Huang, Wei, Chai, Luo, and Wei}]{stdc}
Fan, M.; Lai, S.; Huang, J.; Wei, X.; Chai, Z.; Luo, J.; and Wei, X. 2021.
\newblock Rethinking bisenet for real-time semantic segmentation.
\newblock In \emph{CVPR}, 9716--9725.

\bibitem[{Fang et~al.(2023)Fang, Ma, Song, Mi, and Wang}]{depgraph}
Fang, G.; Ma, X.; Song, M.; Mi, M.~B.; and Wang, X. 2023.
\newblock Depgraph: Towards any structural pruning.
\newblock In \emph{CVPR}, 16091--16101.

\bibitem[{Guo et~al.(2020)Guo, Wang, Li, and Yan}]{guo2020dmcp}
Guo, S.; Wang, Y.; Li, Q.; and Yan, J. 2020.
\newblock Dmcp: Differentiable markov channel pruning for neural networks.
\newblock In \emph{CVPR}, 1539--1547.

\bibitem[{Han, Mao, and Dally(2015)}]{deep_compression}
Han, S.; Mao, H.; and Dally, W.~J. 2015.
\newblock Deep compression: Compressing deep neural networks with pruning, trained quantization and huffman coding.
\newblock \emph{arXiv preprint arXiv:1510.00149}.

\bibitem[{Han et~al.(2015)Han, Pool, Tran, and Dally}]{learning_weights_connections}
Han, S.; Pool, J.; Tran, J.; and Dally, W. 2015.
\newblock Learning both weights and connections for efficient neural network.
\newblock \emph{NeurIPS}, 28.

\bibitem[{He et~al.(2021)He, Wu, Liang, and Lam}]{cap}
He, W.; Wu, M.; Liang, M.; and Lam, S.-K. 2021.
\newblock Cap: Context-aware pruning for semantic segmentation.
\newblock In \emph{WACV}, 960--969.

\bibitem[{He et~al.(2018)He, Kang, Dong, Fu, and Yang}]{sfp}
He, Y.; Kang, G.; Dong, X.; Fu, Y.; and Yang, Y. 2018.
\newblock Soft Filter Pruning for Accelerating Deep Convolutional Neural Networks.
\newblock In \emph{IJCAI}, 2234--2240.

\bibitem[{He et~al.(2019)He, Liu, Wang, Hu, and Yang}]{geometric}
He, Y.; Liu, P.; Wang, Z.; Hu, Z.; and Yang, Y. 2019.
\newblock Filter pruning via geometric median for deep convolutional neural networks acceleration.
\newblock In \emph{CVPR}, 4340--4349.

\bibitem[{Hosseinian, Fontes, and Butenko(2020)}]{hosseinian2020lagrangian}
Hosseinian, S.; Fontes, D.~B.; and Butenko, S. 2020.
\newblock A lagrangian bound on the clique number and an exact algorithm for the maximum edge weight clique problem.
\newblock \emph{INFORMS Journal on Computing}, 32(3): 747--762.

\bibitem[{Hosseinian et~al.(2017)Hosseinian, Fontes, Butenko, Nardelli, Fornari, and Curtarolo}]{hosseinian2017maximum}
Hosseinian, S.; Fontes, D.~B.; Butenko, S.; Nardelli, M.~B.; Fornari, M.; and Curtarolo, S. 2017.
\newblock The maximum edge weight clique problem: formulations and solution approaches.
\newblock \emph{Optimization Methods and Applications: In Honor of Ivan V. Sergienko's 80th Birthday}, 217--237.

\bibitem[{Hou et~al.(2022)Hou, Qin, Sun, Ma, Yuan, Xu, Chen, Jin, Xie, and Kung}]{chex}
Hou, Z.; Qin, M.; Sun, F.; Ma, X.; Yuan, K.; Xu, Y.; Chen, Y.-K.; Jin, R.; Xie, Y.; and Kung, S.-Y. 2022.
\newblock Chex: Channel exploration for cnn model compression.
\newblock In \emph{CVPR}, 12287--12298.

\bibitem[{Huang and Wang(2018)}]{sss}
Huang, Z.; and Wang, N. 2018.
\newblock Data-driven sparse structure selection for deep neural networks.
\newblock In \emph{ECCV}, 304--320.

\bibitem[{Ioffe and Szegedy(2015)}]{bn}
Ioffe, S.; and Szegedy, C. 2015.
\newblock Batch normalization: Accelerating deep network training by reducing internal covariate shift.
\newblock In \emph{ICML}, 448--456.

\bibitem[{Li et~al.(2016)Li, Kadav, Durdanovic, Samet, and Graf}]{Pruning_Filters_for_Efficient_ConvNets}
Li, H.; Kadav, A.; Durdanovic, I.; Samet, H.; and Graf, H. 2016.
\newblock Pruning Filters for Efficient ConvNets.
\newblock \emph{arXiv preprint arXiv: 1608.08710}.

\bibitem[{Li et~al.(2020)Li, Gu, Mayer, Gool, and Timofte}]{hinge}
Li, Y.; Gu, S.; Mayer, C.; Gool, L.~V.; and Timofte, R. 2020.
\newblock Group sparsity: The hinge between filter pruning and decomposition for network compression.
\newblock In \emph{CVPR}, 8018--8027.

\bibitem[{Lin et~al.(2020{\natexlab{a}})Lin, Ji, Wang, Zhang, Zhang, Tian, and Shao}]{hrank}
Lin, M.; Ji, R.; Wang, Y.; Zhang, Y.; Zhang, B.; Tian, Y.; and Shao, L. 2020{\natexlab{a}}.
\newblock Hrank: Filter pruning using high-rank feature map.
\newblock In \emph{CVPR}, 1529--1538.

\bibitem[{Lin et~al.(2020{\natexlab{b}})Lin, Stich, Barba, Dmitriev, and Jaggi}]{dynamic_model_pruning}
Lin, T.; Stich, S.; Barba, L.; Dmitriev, D.; and Jaggi, M. 2020{\natexlab{b}}.
\newblock Dynamic Model Pruning with Feedback.
\newblock \emph{ICLR}.

\bibitem[{Lin et~al.(2014)Lin, Maire, Belongie, Hays, Perona, Ramanan, Doll{\'a}r, and Zitnick}]{coco2017}
Lin, T.-Y.; Maire, M.; Belongie, S.; Hays, J.; Perona, P.; Ramanan, D.; Doll{\'a}r, P.; and Zitnick, C.~L. 2014.
\newblock Microsoft coco: Common objects in context.
\newblock In \emph{ECCV}, 740--755.

\bibitem[{Liu et~al.(2021{\natexlab{a}})Liu, Zhang, Kuang, Zhou, Xue, Wang, Chen, Yang, Liao, and Zhang}]{gfp}
Liu, L.; Zhang, S.; Kuang, Z.; Zhou, A.; Xue, J.-H.; Wang, X.; Chen, Y.; Yang, W.; Liao, Q.; and Zhang, W. 2021{\natexlab{a}}.
\newblock Group fisher pruning for practical network compression.
\newblock In \emph{ICML}, 7021--7032.

\bibitem[{Liu et~al.(2017)Liu, Li, Shen, Huang, Yan, and Zhang}]{networksliming}
Liu, Z.; Li, J.; Shen, Z.; Huang, G.; Yan, S.; and Zhang, C. 2017.
\newblock Learning efficient convolutional networks through network slimming.
\newblock In \emph{CVPR}, 2736--2744.

\bibitem[{Liu et~al.(2021{\natexlab{b}})Liu, Lin, Cao, Hu, Wei, Zhang, Lin, and Guo}]{liu2021swin}
Liu, Z.; Lin, Y.; Cao, Y.; Hu, H.; Wei, Y.; Zhang, Z.; Lin, S.; and Guo, B. 2021{\natexlab{b}}.
\newblock Swin transformer: Hierarchical vision transformer using shifted windows.
\newblock In \emph{ICCV}, 10012--10022.

\bibitem[{Long, Shelhamer, and Darrell(2015)}]{fcn}
Long, J.; Shelhamer, E.; and Darrell, T. 2015.
\newblock Fully convolutional networks for semantic segmentation.
\newblock In \emph{CVPR}, 3431--3440.

\bibitem[{Luo, Wu, and Lin(2017)}]{thinet}
Luo, J.-H.; Wu, J.; and Lin, W. 2017.
\newblock Thinet: A filter level pruning method for deep neural network compression.
\newblock In \emph{CVPR}, 5058--5066.

\bibitem[{Macambira and De~Souza(2000)}]{macambira2000edge}
Macambira, E.~M.; and De~Souza, C.~C. 2000.
\newblock The edge-weighted clique problem: valid inequalities, facets and polyhedral computations.
\newblock \emph{European Journal of Operational Research}, 123(2): 346--371.

\bibitem[{Molchanov et~al.(2019)Molchanov, Mallya, Tyree, Frosio, and Kautz}]{importance_estimation}
Molchanov, P.; Mallya, A.; Tyree, S.; Frosio, I.; and Kautz, J. 2019.
\newblock Importance estimation for neural network pruning.
\newblock In \emph{CVPR}, 11264--11272.

\bibitem[{Ning et~al.(2020)Ning, Zhao, Li, Lei, Wang, and Yang}]{dsa}
Ning, X.; Zhao, T.; Li, W.; Lei, P.; Wang, Y.; and Yang, H. 2020.
\newblock DSA: More Efficient Budgeted Pruning via Differentiable Sparsity Allocation.
\newblock In \emph{ECCV}, 592--607.

\bibitem[{Oh et~al.(2022)Oh, Kim, Baik, Hong, and Lee}]{bnfi}
Oh, J.; Kim, H.; Baik, S.; Hong, C.; and Lee, K.~M. 2022.
\newblock Batch normalization tells you which filter is important.
\newblock In \emph{WACV}, 2645--2654.

\bibitem[{Pan et~al.(2022)Pan, Hong, Sun, and Jia}]{ddrnet}
Pan, H.; Hong, Y.; Sun, W.; and Jia, Y. 2022.
\newblock Deep dual-resolution networks for real-time and accurate semantic segmentation of traffic scenes.
\newblock \emph{T-ITS}, 24(3): 3448--3460.

\bibitem[{Pang et~al.(2019)Pang, Li, Shen, and Shao}]{pang2019towards}
Pang, Y.; Li, Y.; Shen, J.; and Shao, L. 2019.
\newblock Towards bridging semantic gap to improve semantic segmentation.
\newblock In \emph{CVPR}, 4230--4239.

\bibitem[{Peng et~al.(2019)Peng, Wu, Chen, and Huang}]{ccp}
Peng, H.; Wu, J.; Chen, S.; and Huang, J. 2019.
\newblock Collaborative channel pruning for deep networks.
\newblock In \emph{ICML}, 5113--5122.

\bibitem[{Pullan(2008)}]{pullan2008approximating}
Pullan, W. 2008.
\newblock Approximating the maximum vertex/edge weighted clique using local search.
\newblock \emph{Journal of Heuristics}, 14: 117--134.

\bibitem[{Ronneberger, Fischer, and Brox(2015)}]{unet}
Ronneberger, O.; Fischer, P.; and Brox, T. 2015.
\newblock U-net: Convolutional networks for biomedical image segmentation.
\newblock In \emph{Proc. Int. Conf. Med. Image Comput. Comput.-Assist. Intervent.}, 234--241.

\bibitem[{Sp{\"a}th(1985)}]{mewcp}
Sp{\"a}th, H. 1985.
\newblock Heuristically determining cliques of given cardinality and with minimal cost within weighted complete graphs.
\newblock \emph{Zeitschrift f{\"u}r Operations Research}, 29: 125--131.

\bibitem[{Su et~al.(2021)Su, You, Huang, Wang, Qian, Zhang, and Xu}]{cafenet}
Su, X.; You, S.; Huang, T.; Wang, F.; Qian, C.; Zhang, C.; and Xu, C. 2021.
\newblock Locally Free Weight Sharing for Network Width Search.
\newblock \emph{ICLR}.

\bibitem[{Sui et~al.(2021)Sui, Yin, Xie, Phan, Aliari~Zonouz, and Yuan}]{chip}
Sui, Y.; Yin, M.; Xie, Y.; Phan, H.; Aliari~Zonouz, S.; and Yuan, B. 2021.
\newblock Chip: Channel independence-based pruning for compact neural networks.
\newblock \emph{NeurIPS}, 34: 24604--24616.

\bibitem[{Wan et~al.(2023)Wan, Huang, Lu, Yu, and Zhang}]{wan2023seaformer}
Wan, Q.; Huang, Z.; Lu, J.; Yu, G.; and Zhang, L. 2023.
\newblock Seaformer: Squeeze-enhanced axial transformer for mobile semantic segmentation.
\newblock \emph{arXiv preprint arXiv:2301.13156}.

\bibitem[{Wang et~al.(2020)Wang, Qin, Zhang, and Fu}]{greg}
Wang, H.; Qin, C.; Zhang, Y.; and Fu, Y. 2020.
\newblock Neural pruning via growing regularization.
\newblock \emph{arXiv preprint arXiv:2012.09243}.

\bibitem[{Wang et~al.(2022)Wang, Gou, Wu, Feng, Han, Ding, and Wang}]{wang2022rtformer}
Wang, J.; Gou, C.; Wu, Q.; Feng, H.; Han, J.; Ding, E.; and Wang, J. 2022.
\newblock RTFormer: Efficient design for real-time semantic segmentation with transformer.
\newblock \emph{NeurIPS}, 35: 7423--7436.

\bibitem[{Wang et~al.(2024)Wang, Xie, Wang, Xu, and Jin}]{dcfp}
Wang, Z.; Xie, H.; Wang, Y.; Xu, H.; and Jin, G. 2024.
\newblock DCFP: Distribution Calibrated Filter Pruning for Lightweight and Accurate Long-tail Semantic Segmentation.
\newblock \emph{TCSVT}, 34(7): 6063--6076.

\bibitem[{Xu et~al.(2018)Xu, Wang, Zhou, Lin, and Xiong}]{single_multilevel_quantization}
Xu, Y.; Wang, Y.; Zhou, A.; Lin, W.; and Xiong, H. 2018.
\newblock Deep neural network compression with single and multiple level quantization.
\newblock In \emph{AAAI}, volume~32.

\bibitem[{Yang et~al.(2019)Yang, Wang, Liu, Chen, Xu, Shi, Xu, and Xu}]{legonet}
Yang, Z.; Wang, Y.; Liu, C.; Chen, H.; Xu, C.; Shi, B.; Xu, C.; and Xu, C. 2019.
\newblock Legonet: Efficient convolutional neural networks with lego filters.
\newblock In \emph{ICML}, 7005--7014.

\bibitem[{Yu et~al.(2021)Yu, Gao, Wang, Yu, Shen, and Sang}]{yu2021bisenet}
Yu, C.; Gao, C.; Wang, J.; Yu, G.; Shen, C.; and Sang, N. 2021.
\newblock Bisenet v2: Bilateral network with guided aggregation for real-time semantic segmentation.
\newblock \emph{IJCV}, 129: 3051--3068.

\bibitem[{Yu et~al.(2018)Yu, Wang, Peng, Gao, Yu, and Sang}]{yu2018bisenet}
Yu, C.; Wang, J.; Peng, C.; Gao, C.; Yu, G.; and Sang, N. 2018.
\newblock Bisenet: Bilateral segmentation network for real-time semantic segmentation.
\newblock In \emph{ECCV}, 325--341.

\bibitem[{Yu et~al.(2017)Yu, Liu, Wang, and Tao}]{on_compressing}
Yu, X.; Liu, T.; Wang, X.; and Tao, D. 2017.
\newblock On compressing deep models by low rank and sparse decomposition.
\newblock In \emph{CVPR}, 7370--7379.

\bibitem[{Zhao et~al.(2017)Zhao, Shi, Qi, Wang, and Jia}]{psp}
Zhao, H.; Shi, J.; Qi, X.; Wang, X.; and Jia, J. 2017.
\newblock Pyramid scene parsing network.
\newblock In \emph{CVPR}, 2881--2890.

\bibitem[{Zhou et~al.(2017)Zhou, Zhao, Puig, Fidler, Barriuso, and Torralba}]{ade}
Zhou, B.; Zhao, H.; Puig, X.; Fidler, S.; Barriuso, A.; and Torralba, A. 2017.
\newblock Scene parsing through ade20k dataset.
\newblock In \emph{CVPR}, 633--641.

\bibitem[{Zhuang et~al.(2018)Zhuang, Tan, Zhuang, Liu, Guo, Wu, Huang, and Zhu}]{Discrimination-aware}
Zhuang, Z.; Tan, M.; Zhuang, B.; Liu, J.; Guo, Y.; Wu, Q.; Huang, J.; and Zhu, J. 2018.
\newblock Discrimination-aware channel pruning for deep neural networks.
\newblock \emph{NeurIPS}, 31.

\end{thebibliography}

\newpage

\appendix

\section{A.~Comparison with other Pruning Methods}

To better show the differences between our SIRFP and other popular pruning methods, we compare SIRFP with some of them in detail.

\subsection{Comparison with FPGM}
As a popular and classic representative method of property-based pruning methods, FPGM~\cite{geometric} prunes the filters(channels) according to their similarity with the geometric median filters, only based on the filter parameters.
Taking FPGM as an example, we compare our method with it to clearly show the differences between SIRFP and property-based pruning methods.

First, FPGM prunes the top-k channels most similar to the geometric median of all other channels in the original un-pruned network.
In contrast, our method aims to minimize redundancy in the pruned network.
To be more specific, if we rethink the optimization goal from the graph theory perspective, FPGM aims to prune the vertices that have the bigger sum of edge weights in the original graph. On the contrary, our method aims at minimizing the sum of the edge-weight of all vertices in the pruned graph.
A simple example will help our readers to understand this difference.
Let's assume two channels are extremely similar to each other but very different from the others. FPGM will prune both of them since their mutual similarity is so large that makes them the top two ones with the largest sum of similarity. However, our method may prune one of them at first but keep the other one since keeping one of them in the pruned network may not cause a large similarity, as long as the other one is not kept in the pruned network together with it at the same time.

Second, our method adopts the data-driven feature redundancy, which can capture the spatial overlap between the feature maps. 
Nevertheless, FPGM calculates the similarity only based on the weight of filters and focuses on the spatial-agnostic semantic information.
This difference makes our method more adaptable for local-sensitive tasks, such as semantic segmentation and object detection.

\subsection{Comparison with CAP}
As a segmentation-specialized method, CAP~\cite{cap} is the pioneering work designed for semantic segmentation.
CAP chooses to utilize the similarity between pooled~(or sub-sampled) and flattened features to calculate the similarity between features for the quantification of context.
Then the quantified context is used for the generation of regularization terms which are then used to punish channels with less context.

There are several major differences between SIRFP and CAP.
The most important one is that SIRFP directly uses the un-pooled feature maps to calculate redundancy, which can preserve rich spatial information in the features and embed the detailed spatial difference into the redundancy.
Instead, CAP adopts multiple pooling and sub-sampling in the context quantification process.
This difference in the feature resolution leads to a big impact on the pruning performance, which is also demonstrated by the results in Table~\ref{tab:resolution} of this appendix.
The large performance gap between our SIRFP and CAP shown in the Figure~1 of the main text also corroborates our claim that the pruning metric for semantic segmentation should incorporate the difference in spatial information which is vital for segmentation networks.
Secondly, our method does not intervene in the learning of the pre-trained network. 
As the performance of the pruned network is closely related to that of the pre-trained network, the adoption of the regularization terms of CAP has an influence on the mIoU results of the pruned network.
However, SIRFP does not change the gradients during the pre-training stage, leaving the pre-trained networks to learn to segment merely under the supervision of segmentation loss, which could be one of the reasons for the significant advantage of our SIRFP over CAP, as shown in Figure 1 in our submitted main text.

\begin{figure*}[h!]
\centering
\begin{center}
\begin{tabular}{ccccc}

\includegraphics[width=0.19\linewidth]{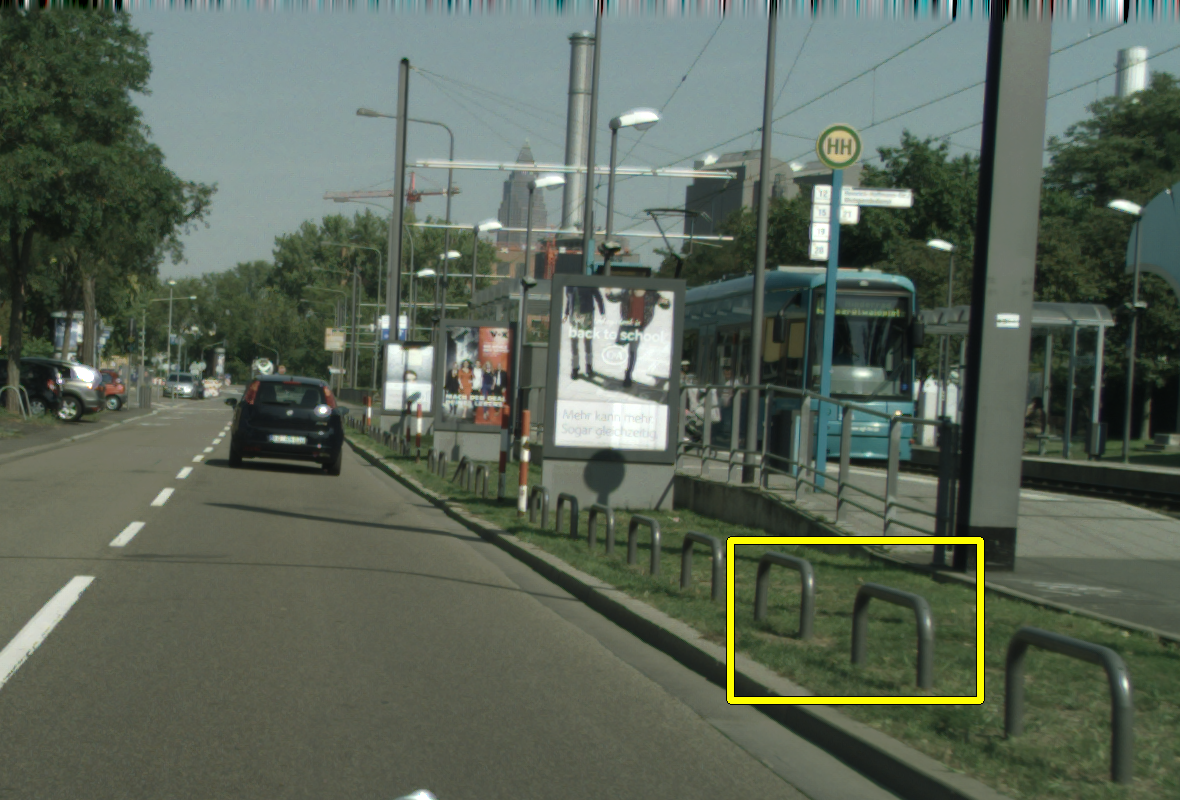}
\hspace{-12pt}
& \includegraphics[width=0.19\linewidth]{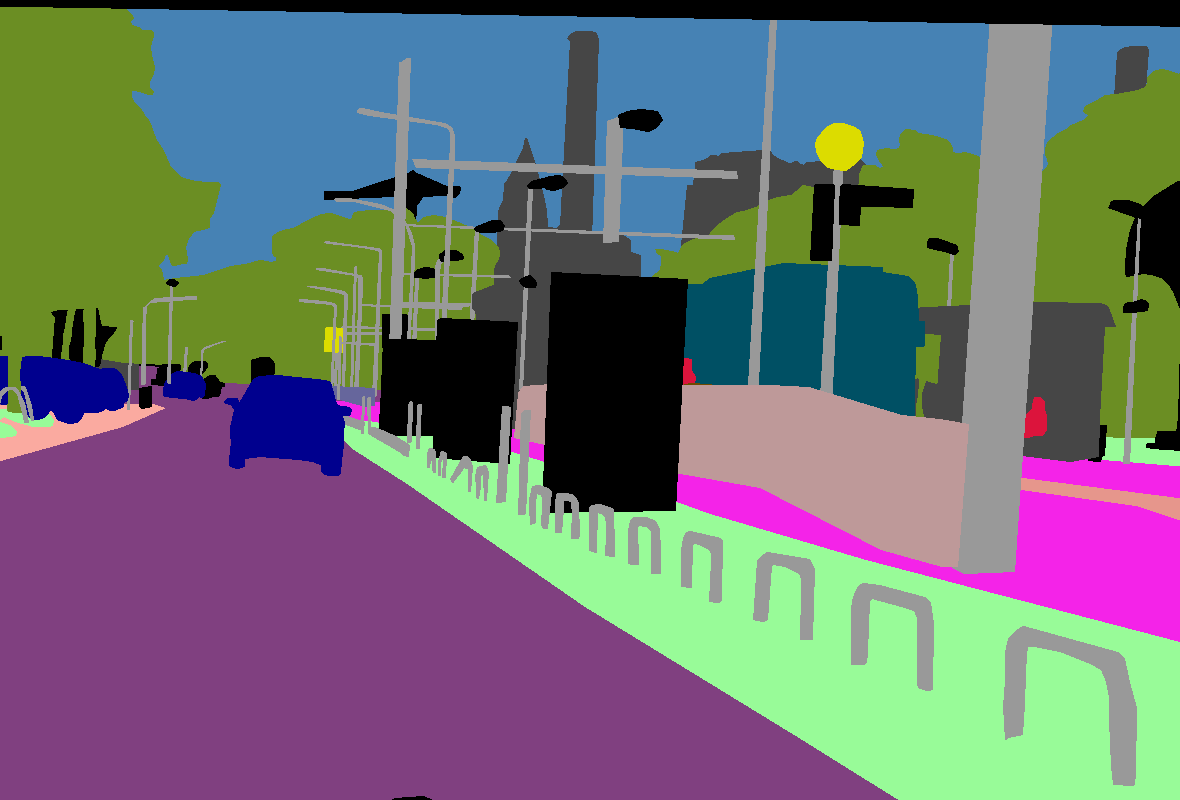}
\hspace{-12pt}
& \includegraphics[width=0.19\linewidth]{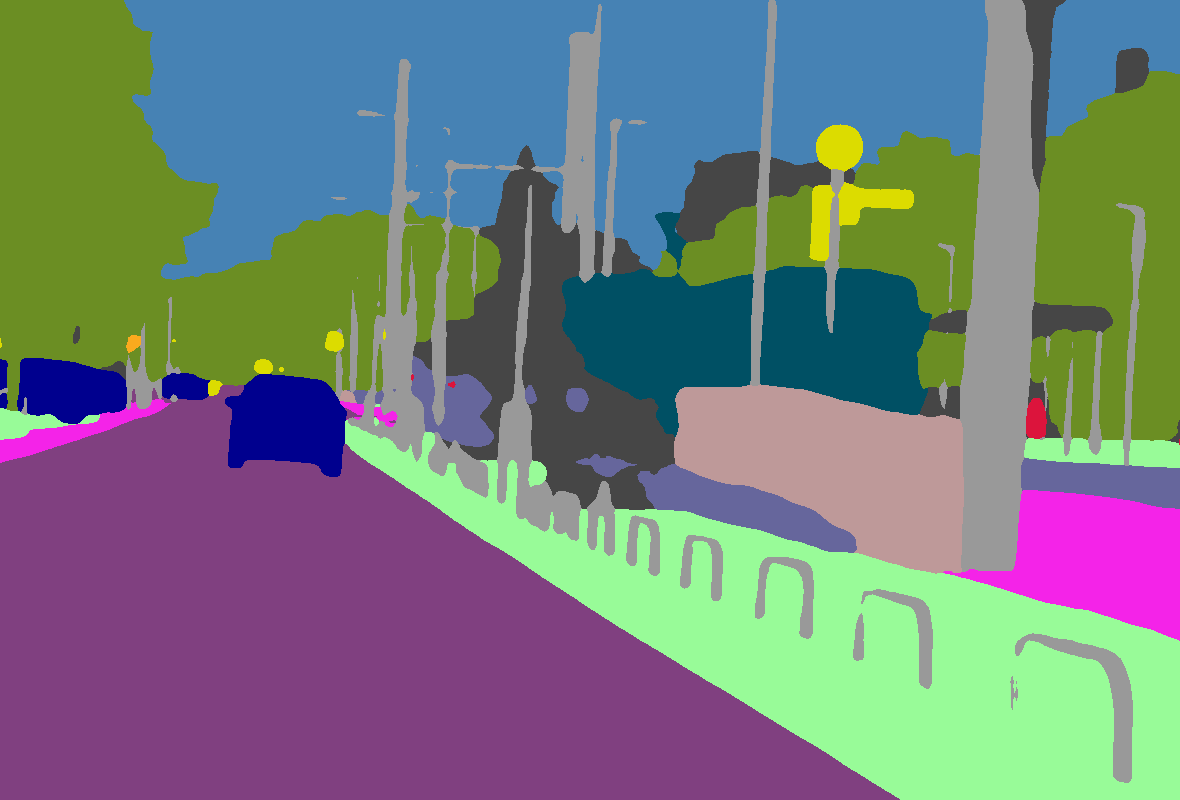}
\hspace{-12pt}
& \includegraphics[width=0.19\linewidth]{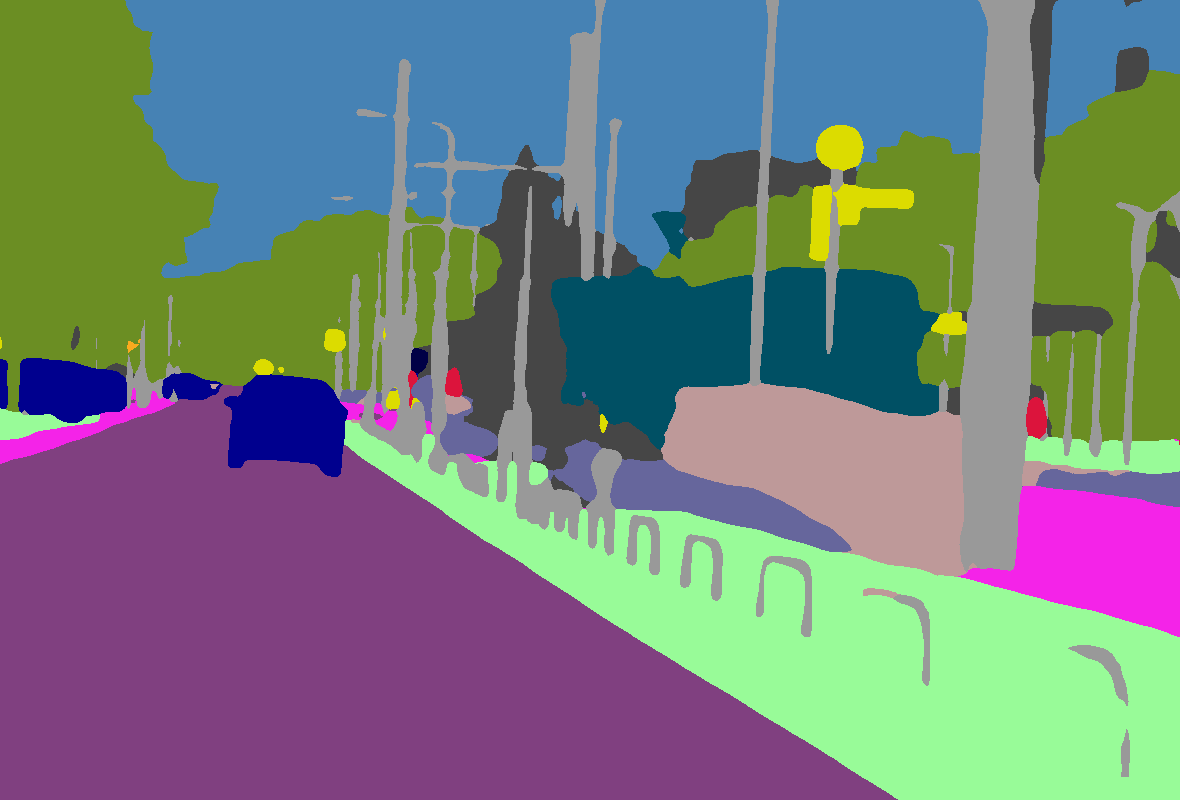}
\hspace{-12pt}
& \includegraphics[width=0.19\linewidth]{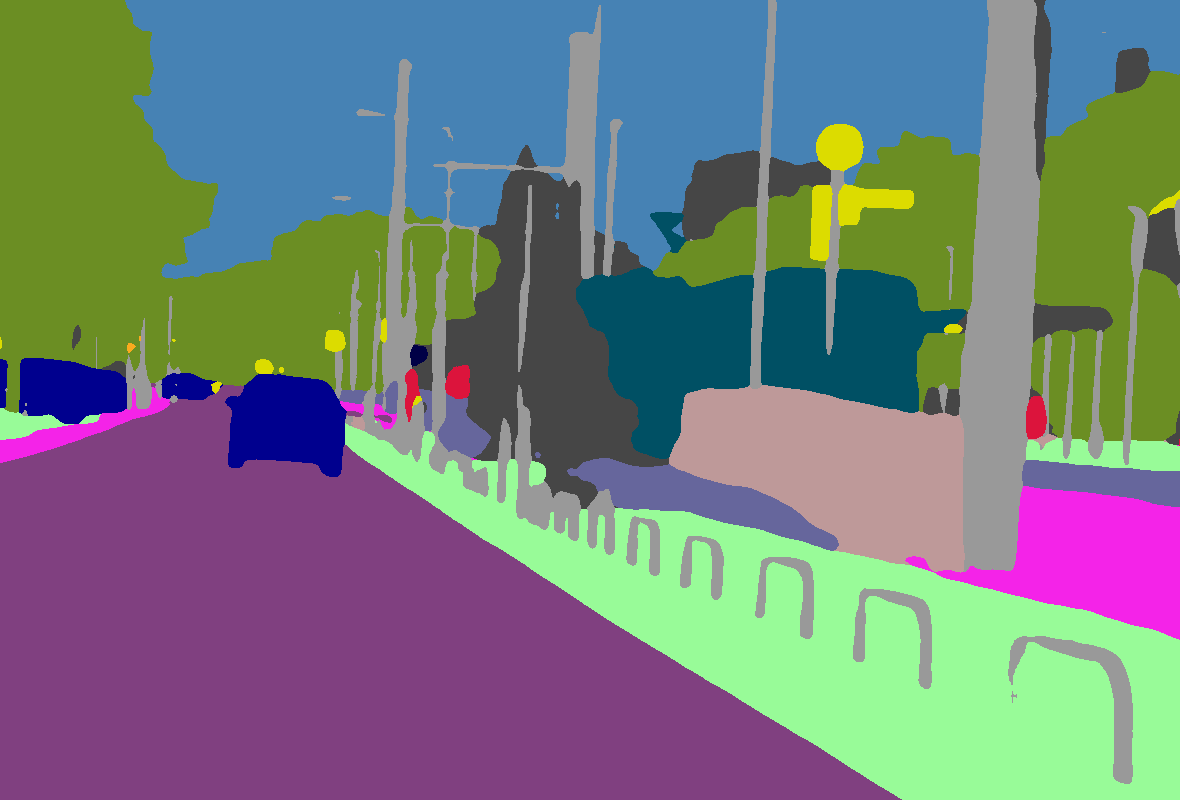}
\\

\includegraphics[width=0.19\linewidth]{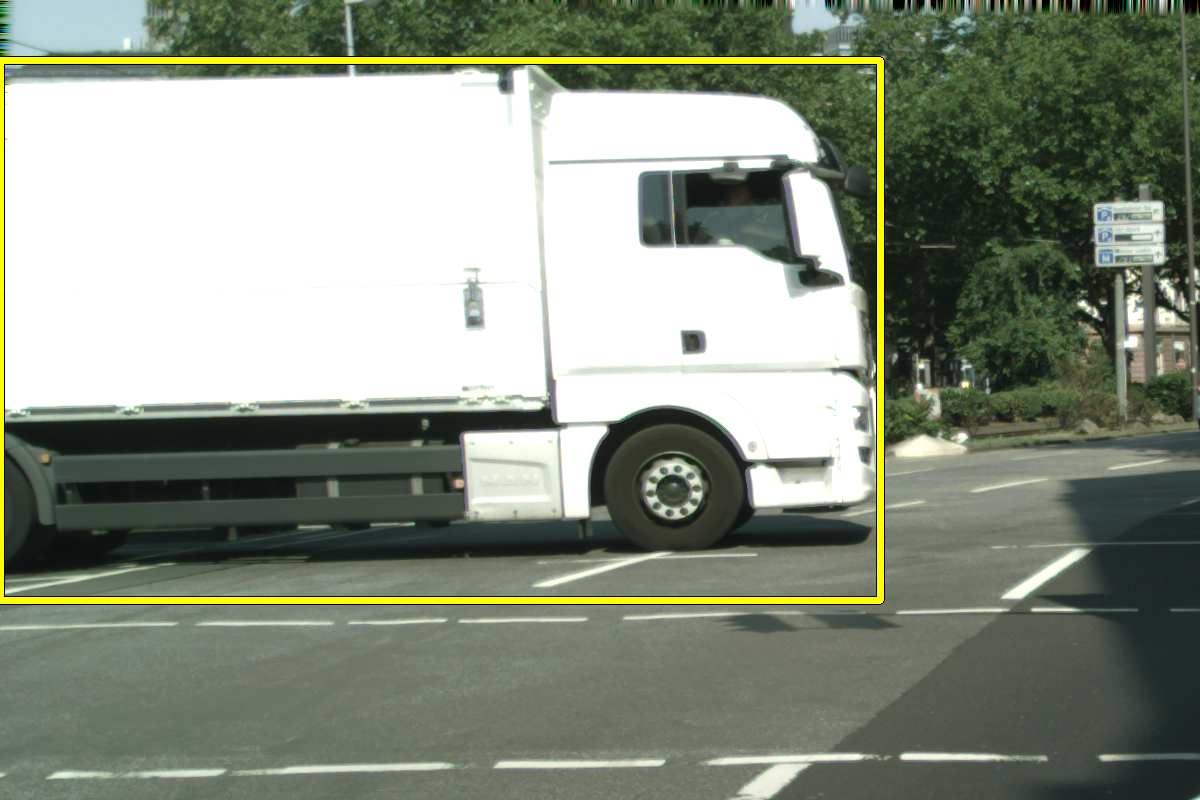}
\hspace{-12pt}
& \includegraphics[width=0.19\linewidth]{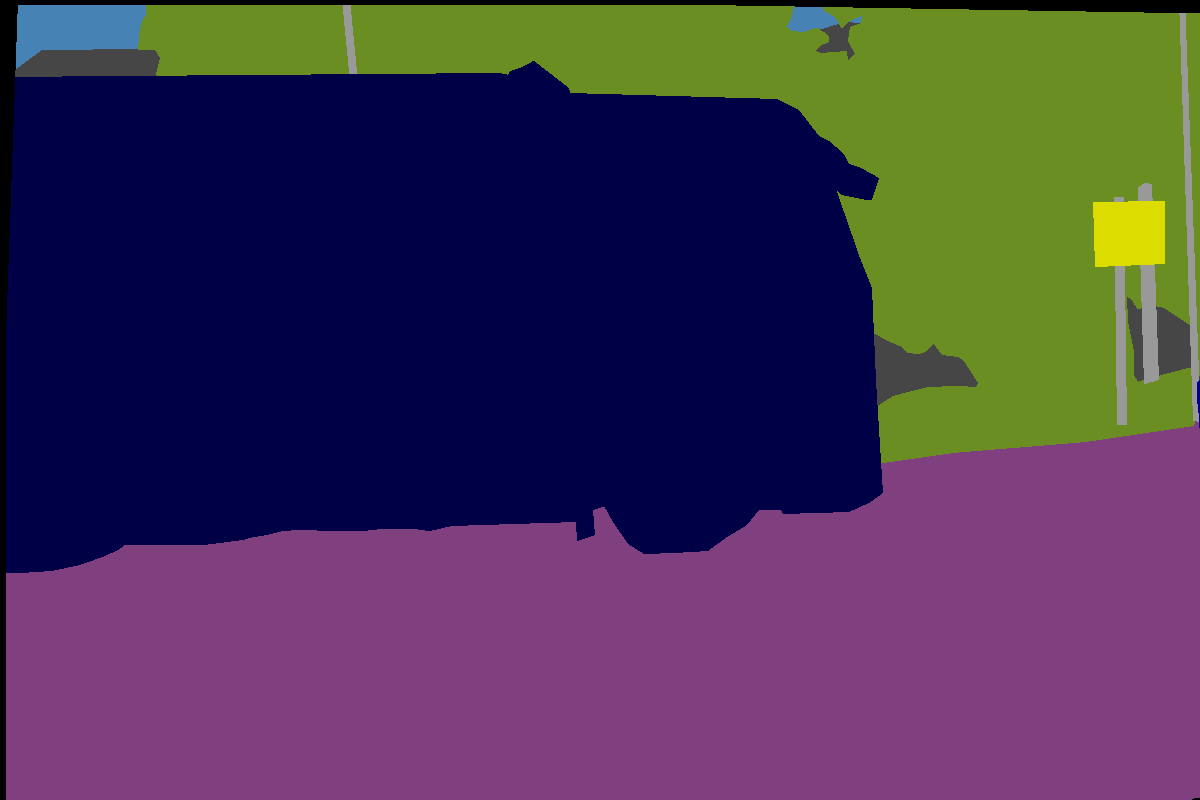}
\hspace{-12pt}
& \includegraphics[width=0.19\linewidth]{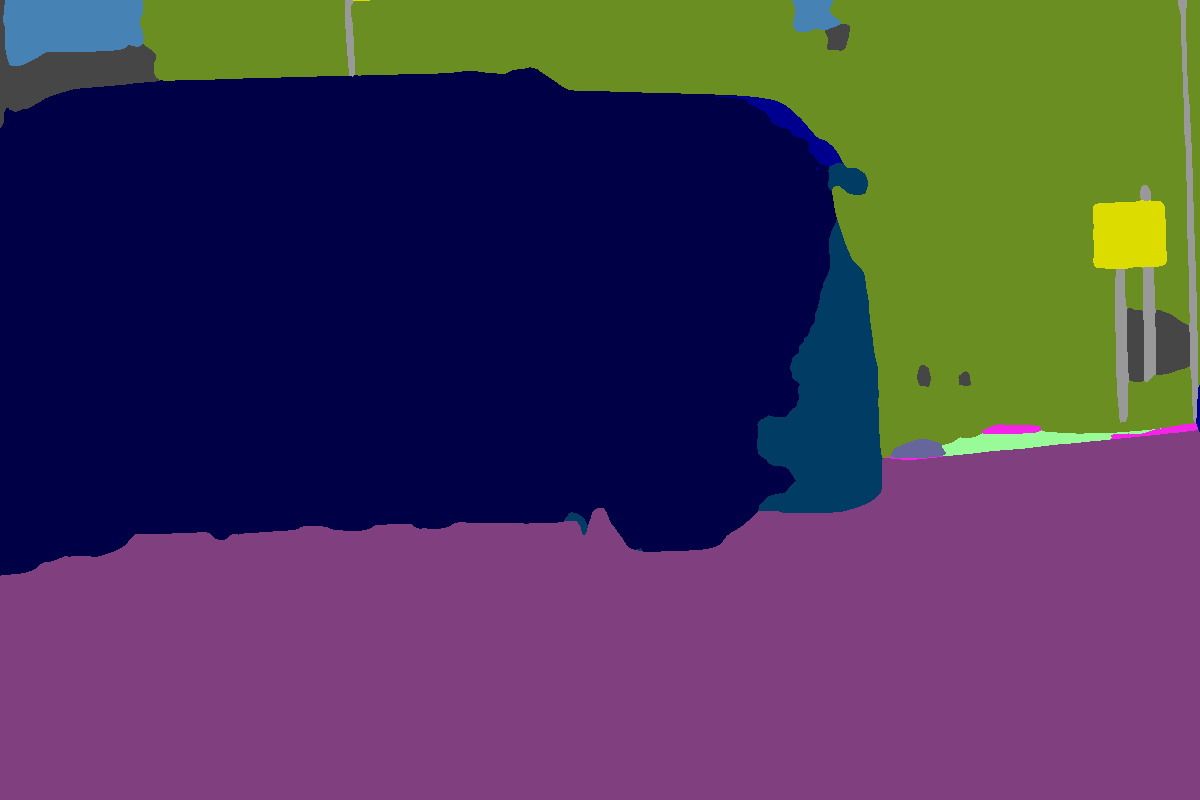}
\hspace{-12pt}
& \includegraphics[width=0.19\linewidth]{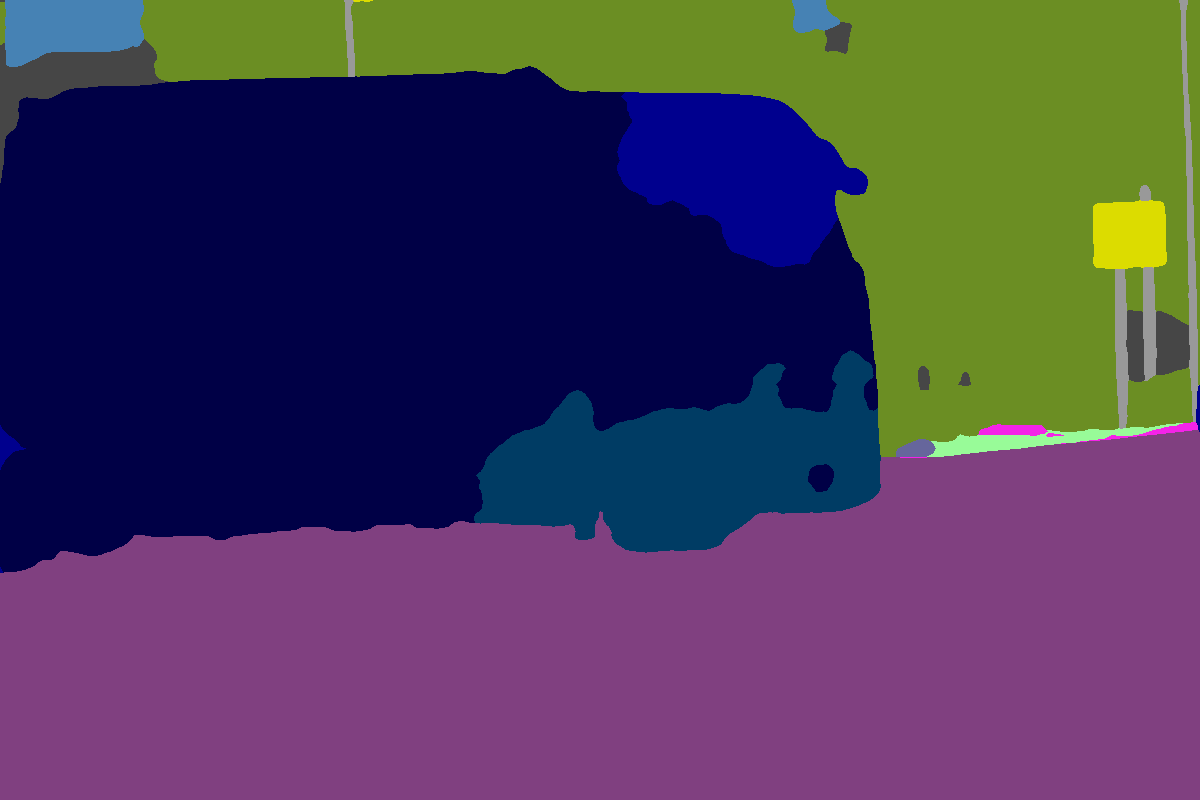}
\hspace{-12pt}
& \includegraphics[width=0.19\linewidth]{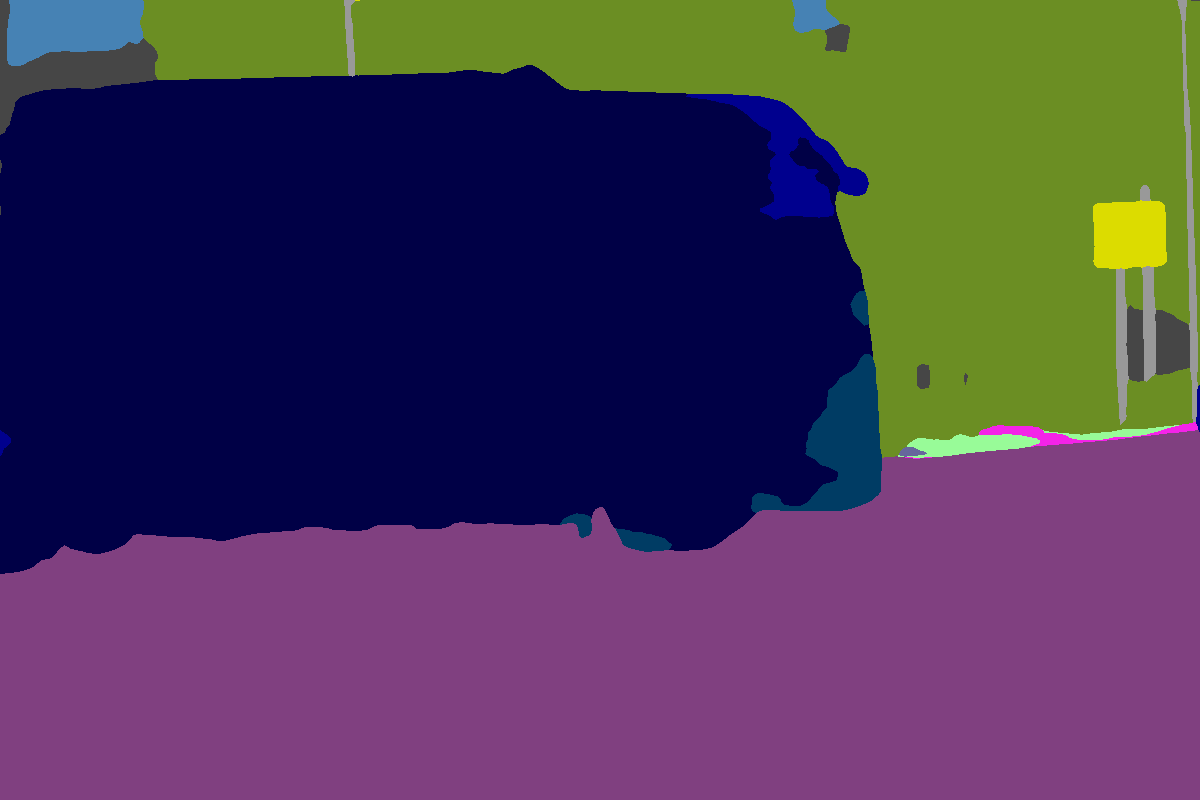}
\\

\includegraphics[width=0.19\linewidth]{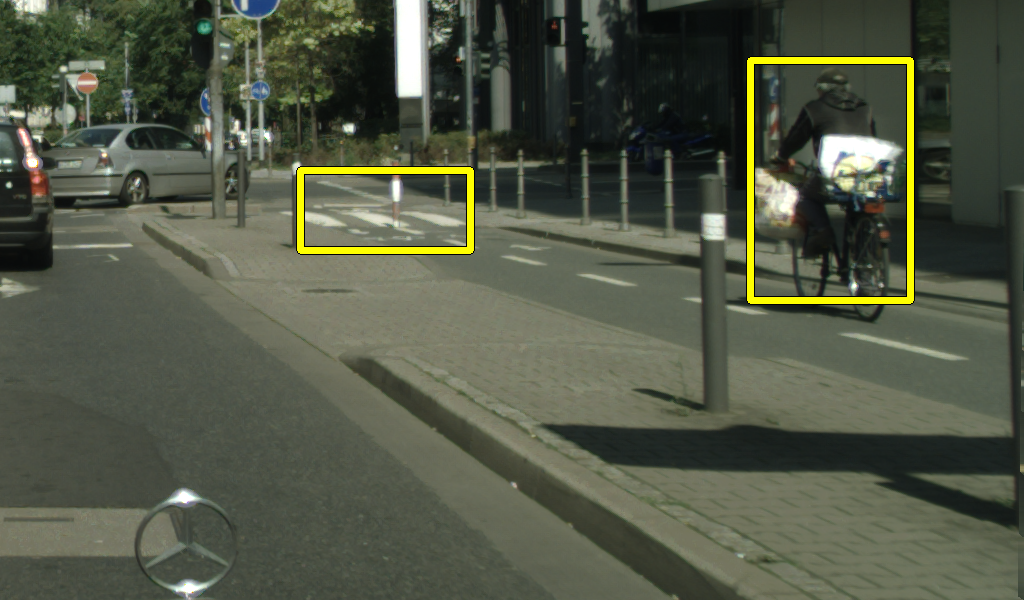}
\hspace{-12pt}
& \includegraphics[width=0.19\linewidth]{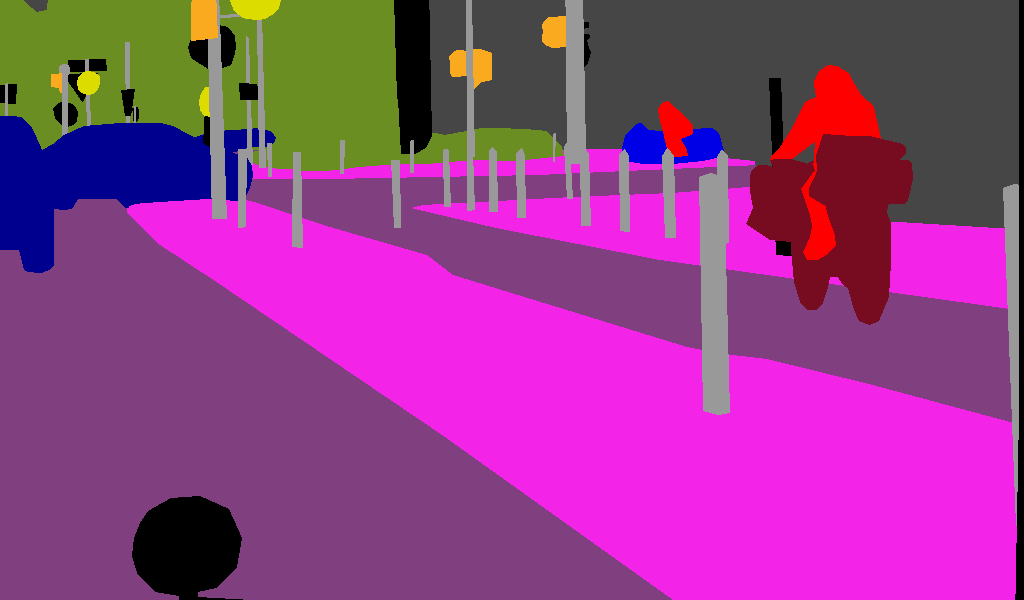}
\hspace{-12pt}
& \includegraphics[width=0.19\linewidth]{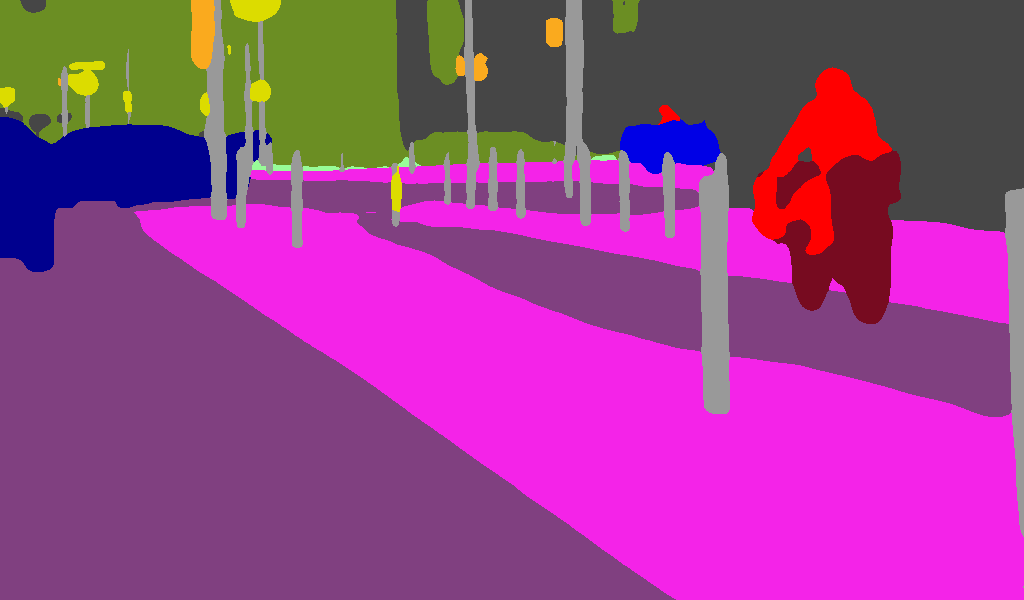}
\hspace{-12pt}
& \includegraphics[width=0.19\linewidth]{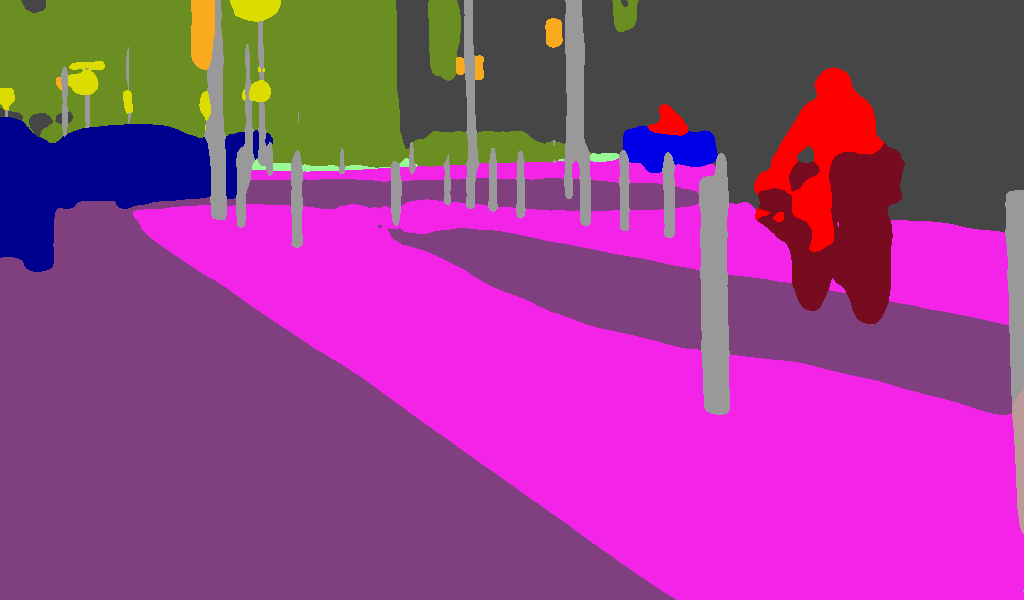}
\hspace{-12pt}
& \includegraphics[width=0.19\linewidth]{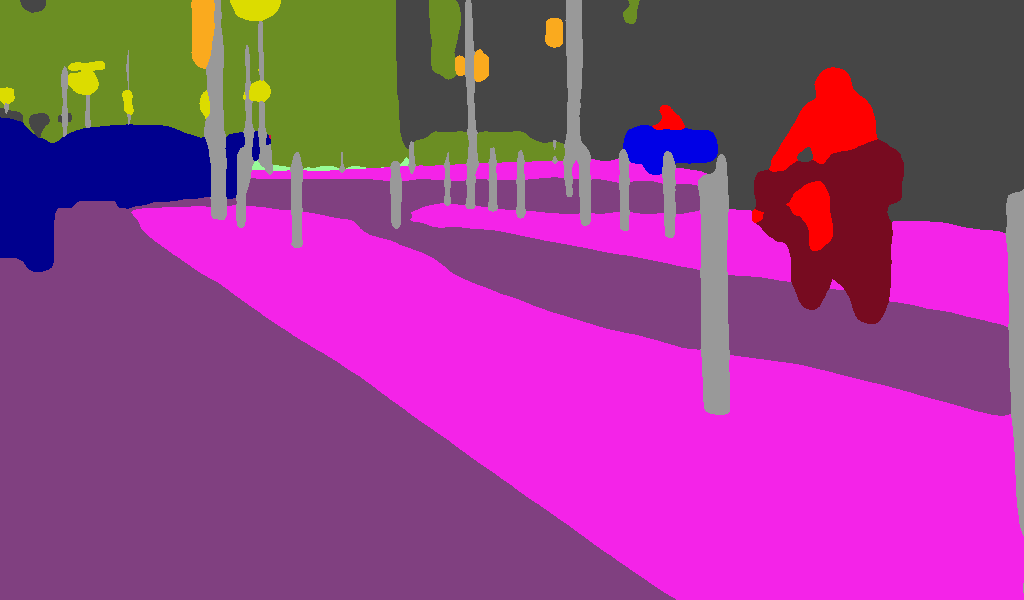}
\\

\includegraphics[width=0.19\linewidth]{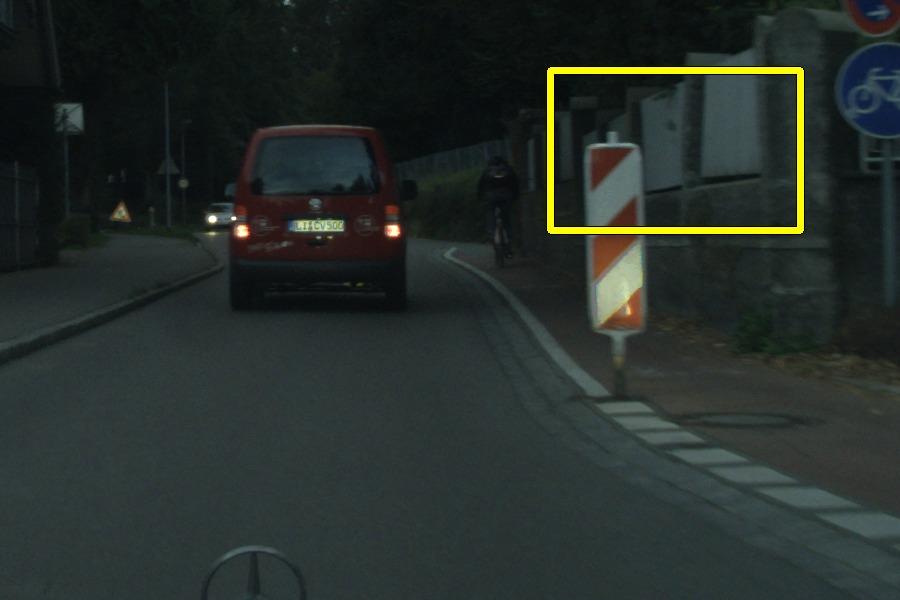}
\hspace{-12pt}
& \includegraphics[width=0.19\linewidth]{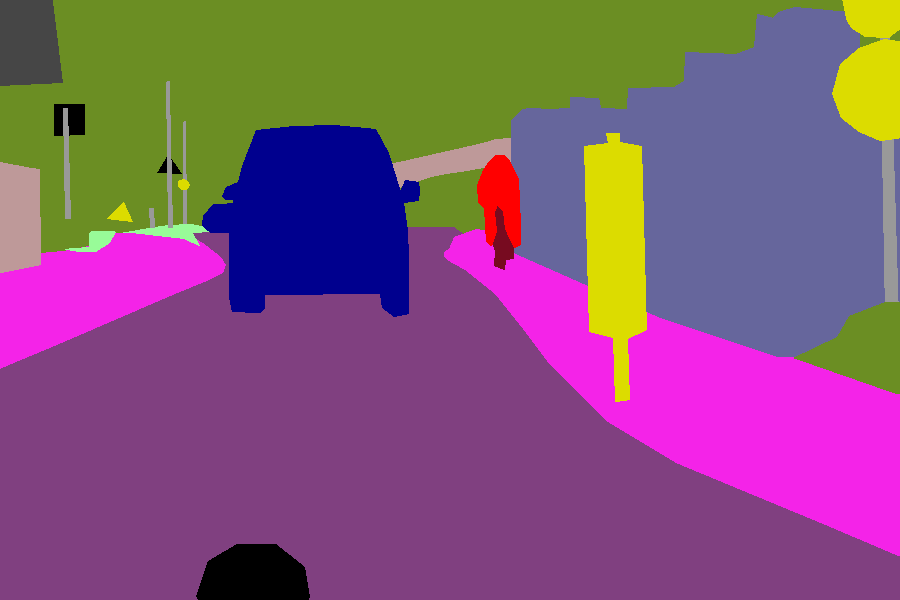}
\hspace{-12pt}
& \includegraphics[width=0.19\linewidth]{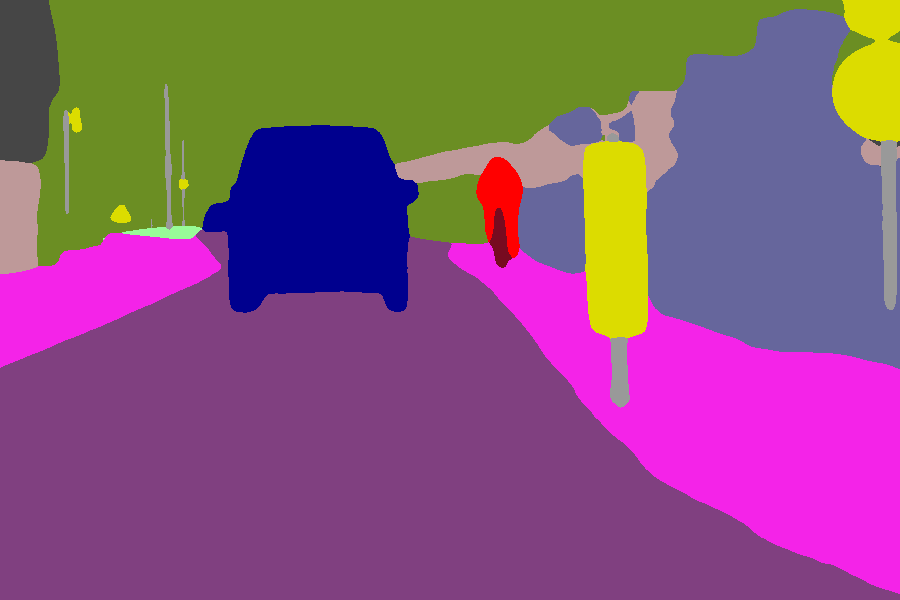}
\hspace{-12pt}
& \includegraphics[width=0.19\linewidth]{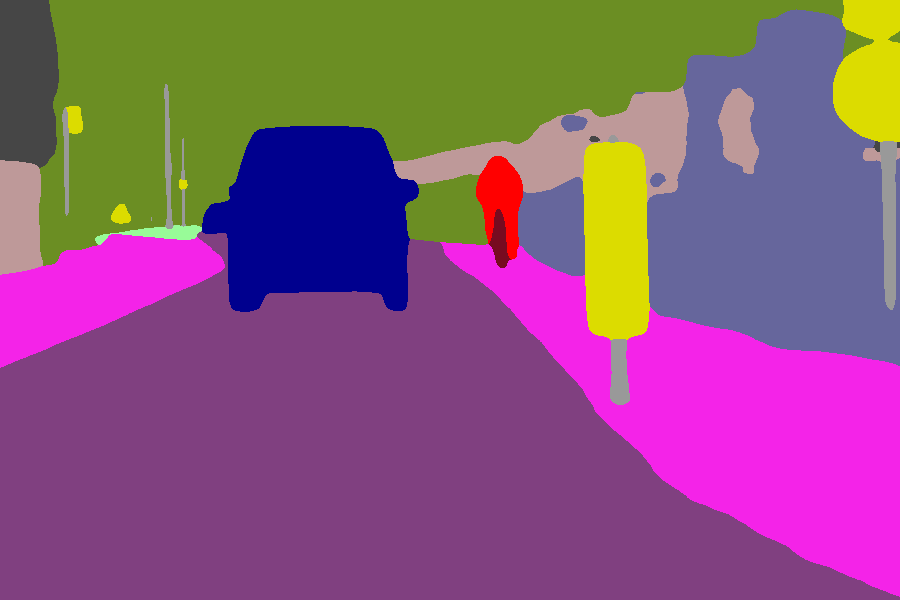}
\hspace{-12pt}
& \includegraphics[width=0.19\linewidth]{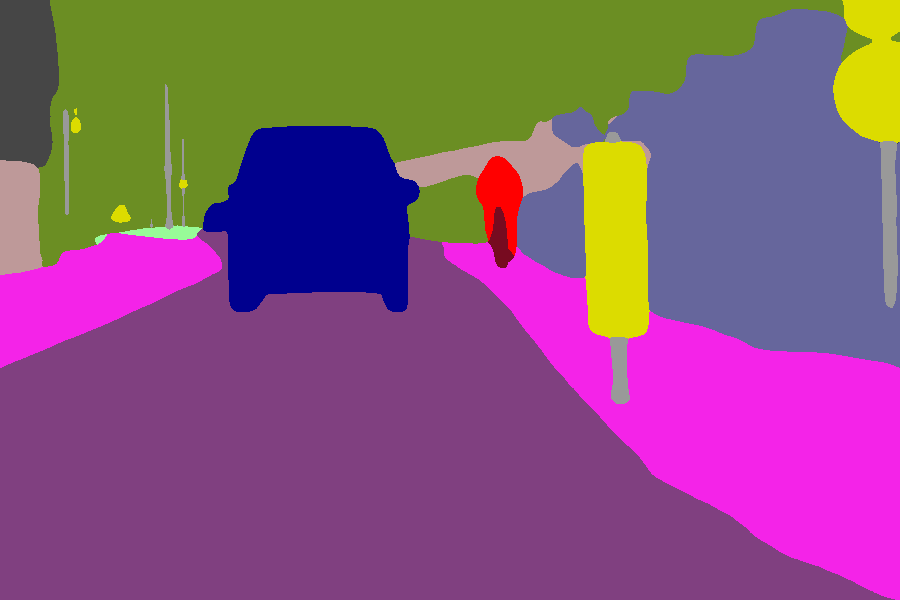}
\\
Image & Ground Truth & Unpruned & DCFP-60\% & SIRFP-60\%~(Ours) \\
\end{tabular}
\end{center}
\caption{
Visualization comparison with DCFP~\cite{dcfp} on Cityscapes validation set with 60\% FLOPs reduction of Deeplabv3-ResNet50.
Black in GT denotes the pixels that should be ignored for evaluation.
We crop the images and use the yellow box in the original images to highlight the difference in predictions between DCFP and SIRFP.
}
\label{fig:city}
\end{figure*}

\section{B.~Visualization Results}

\begin{figure*}[h!]
\centering
\begin{center}
\begin{tabular}{ccccc}

\includegraphics[width=0.19\linewidth]{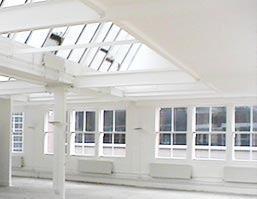}
\hspace{-12pt}
& \includegraphics[width=0.19\linewidth]{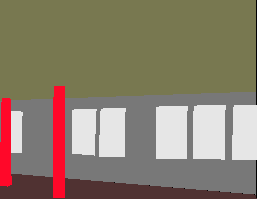}
\hspace{-12pt}
& \includegraphics[width=0.19\linewidth]{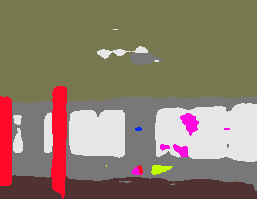}
\hspace{-12pt}
& \includegraphics[width=0.19\linewidth]{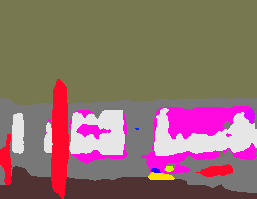}
\hspace{-12pt}
& \includegraphics[width=0.19\linewidth]{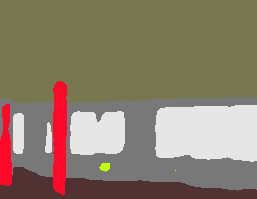}
\\

\includegraphics[width=0.19\linewidth]{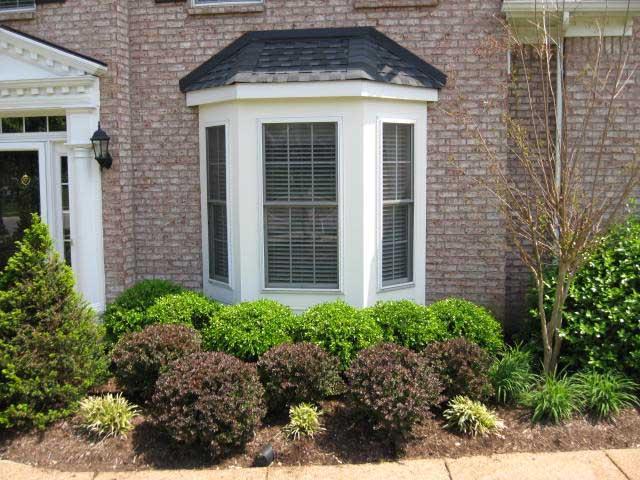}
\hspace{-12pt}
& \includegraphics[width=0.19\linewidth]{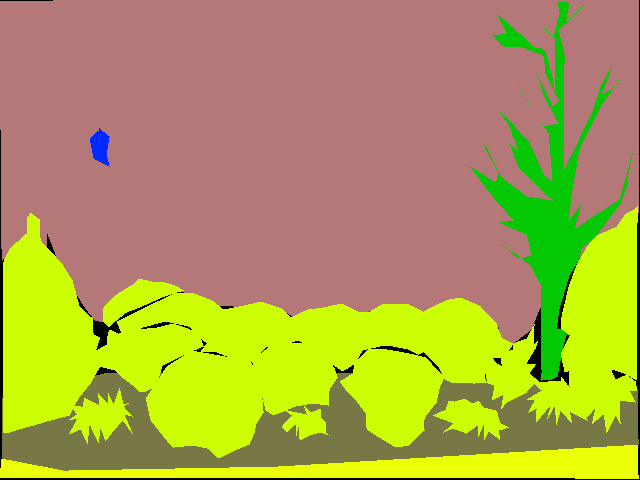}
\hspace{-12pt}
& \includegraphics[width=0.19\linewidth]{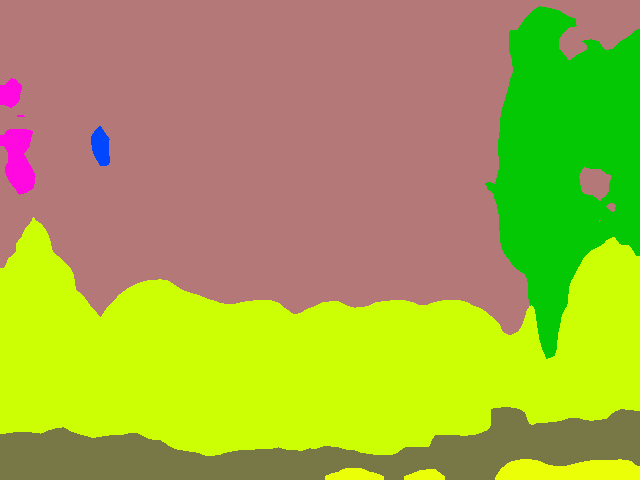}
\hspace{-12pt}
& \includegraphics[width=0.19\linewidth]{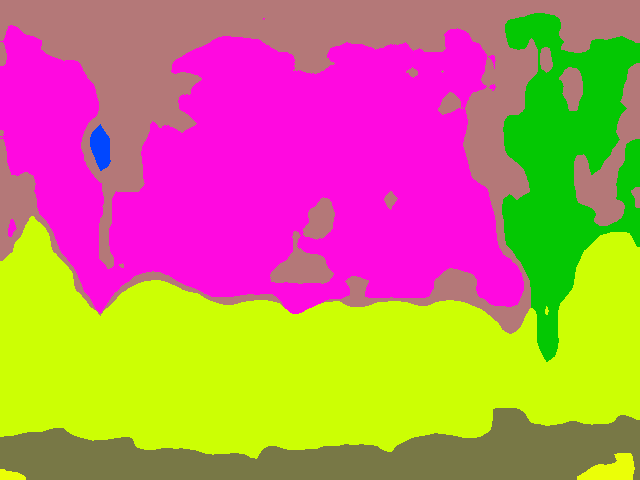}
\hspace{-12pt}
& \includegraphics[width=0.19\linewidth]{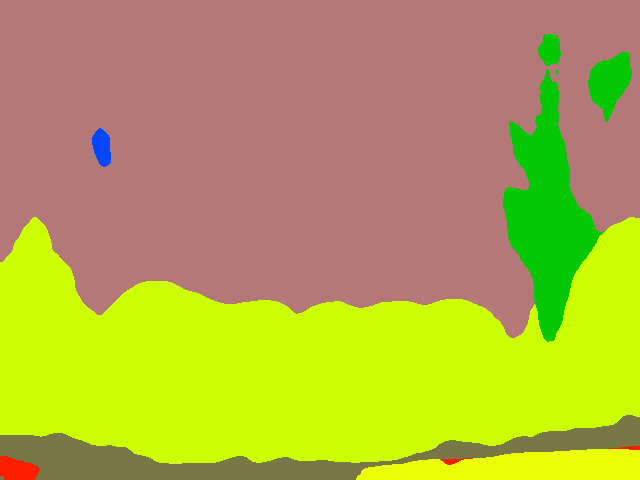}
\\

\includegraphics[width=0.19\linewidth]{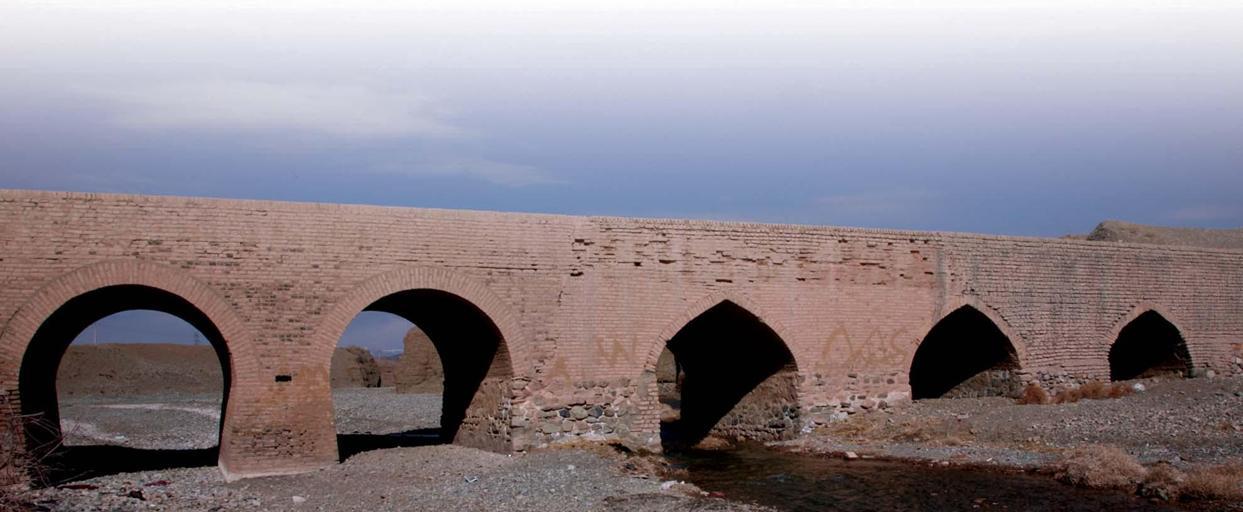}
\hspace{-12pt}
& \includegraphics[width=0.19\linewidth]{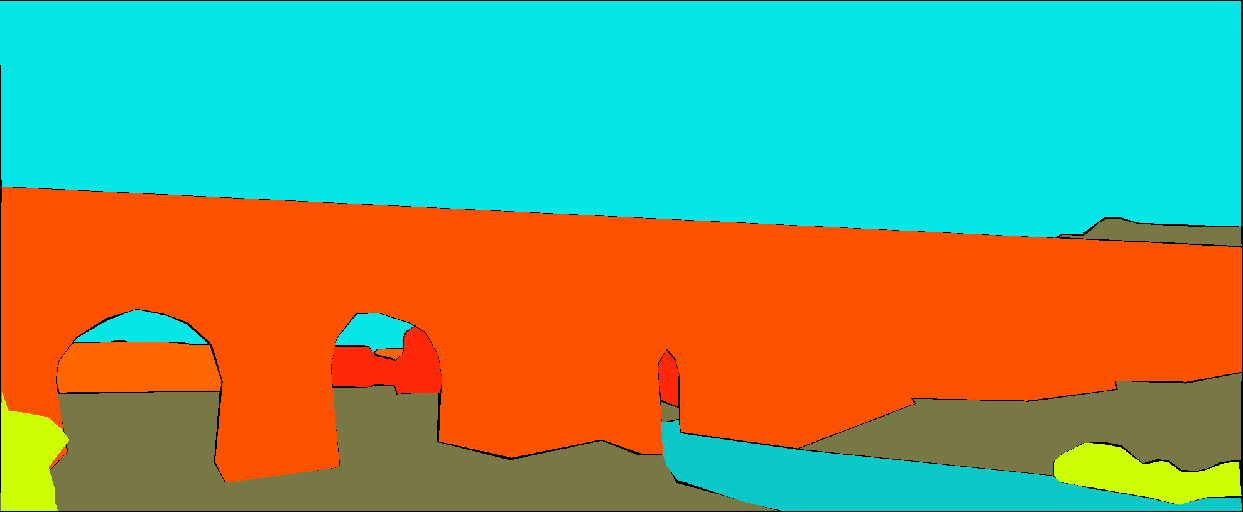}
\hspace{-12pt}
& \includegraphics[width=0.19\linewidth]{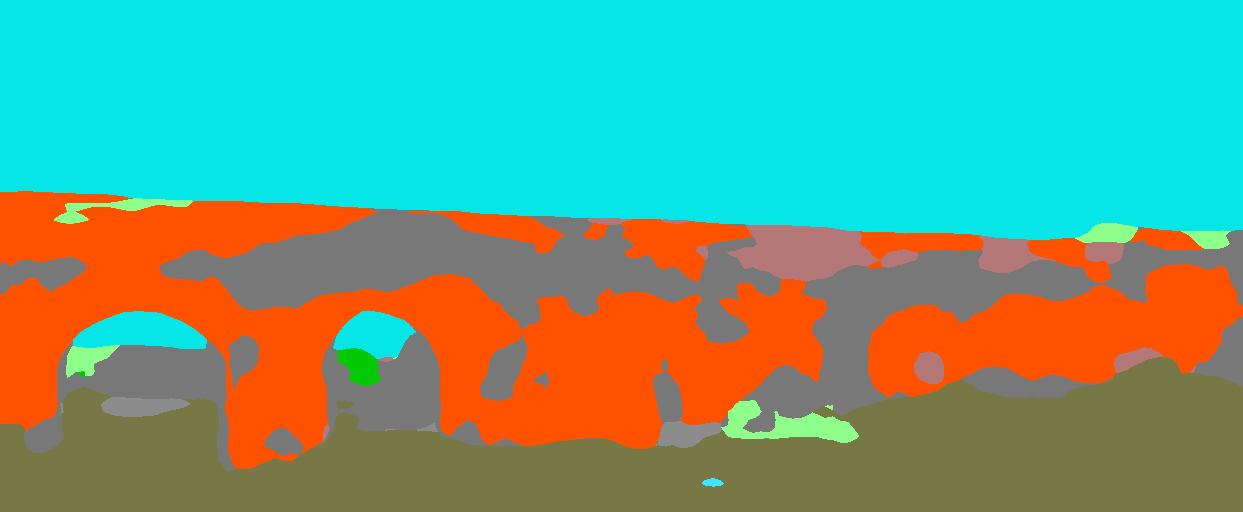}
\hspace{-12pt}
& \includegraphics[width=0.19\linewidth]{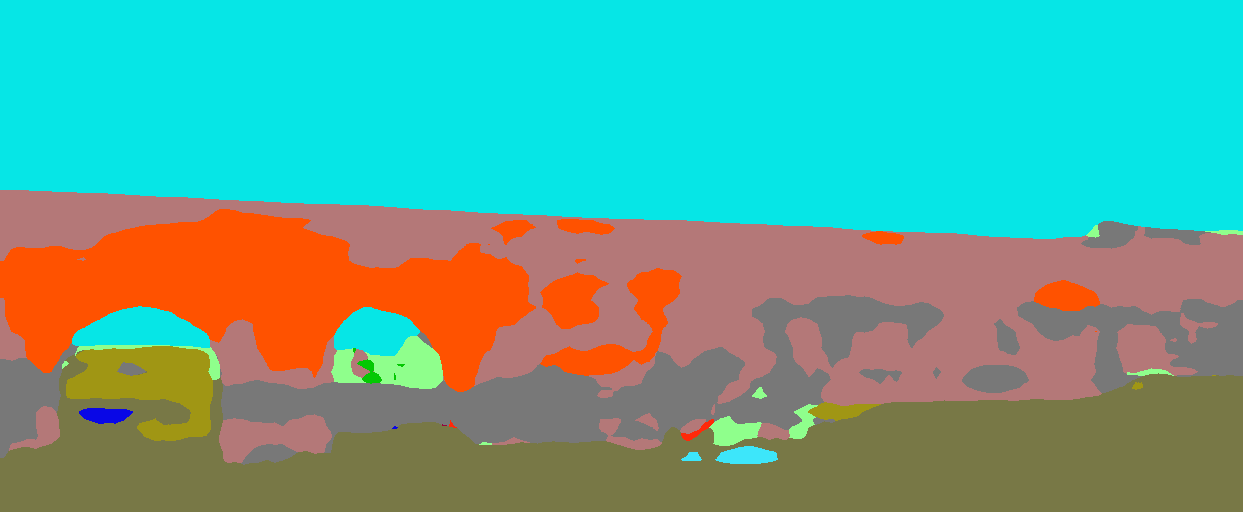}
\hspace{-12pt}
& \includegraphics[width=0.19\linewidth]{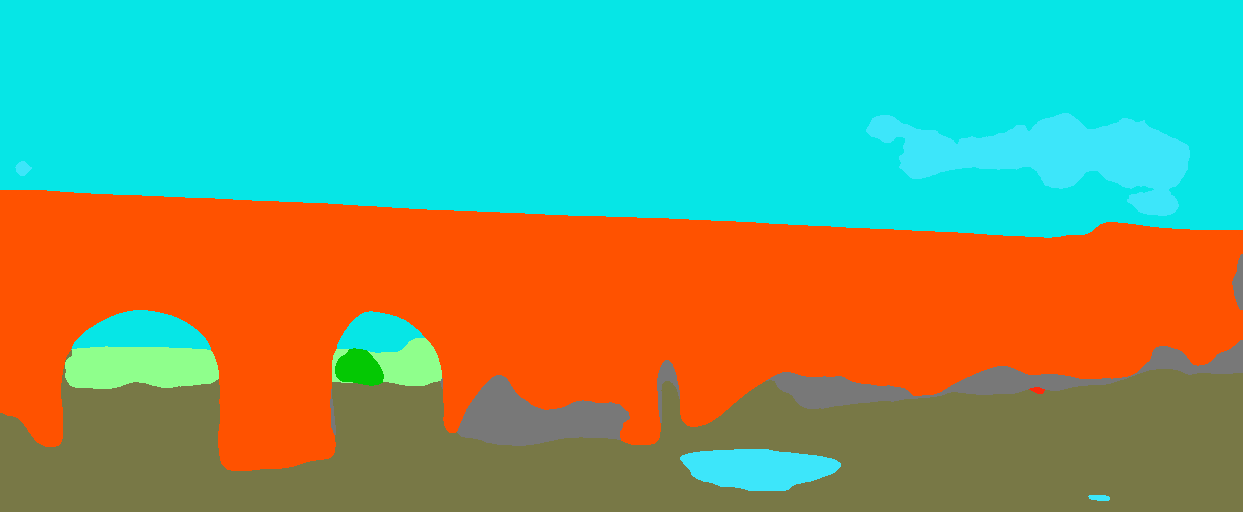}
\\

\includegraphics[width=0.19\linewidth]{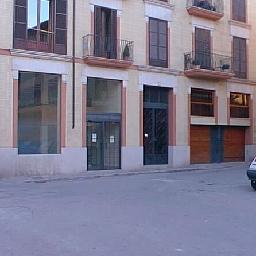}
\hspace{-12pt}
& \includegraphics[width=0.19\linewidth]{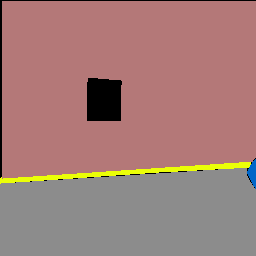}
\hspace{-12pt}
& \includegraphics[width=0.19\linewidth]{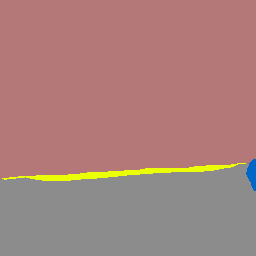}
\hspace{-12pt}
& \includegraphics[width=0.19\linewidth]{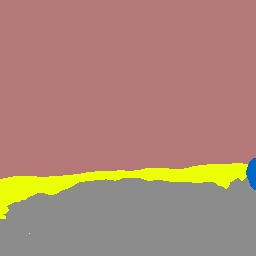}
\hspace{-12pt}
& \includegraphics[width=0.19\linewidth]{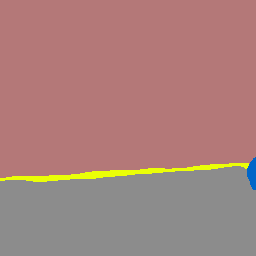}
\\

\includegraphics[width=0.19\linewidth]{images/ade/image/ADE_val_00000314.jpg}
\hspace{-12pt}
& \includegraphics[width=0.19\linewidth]{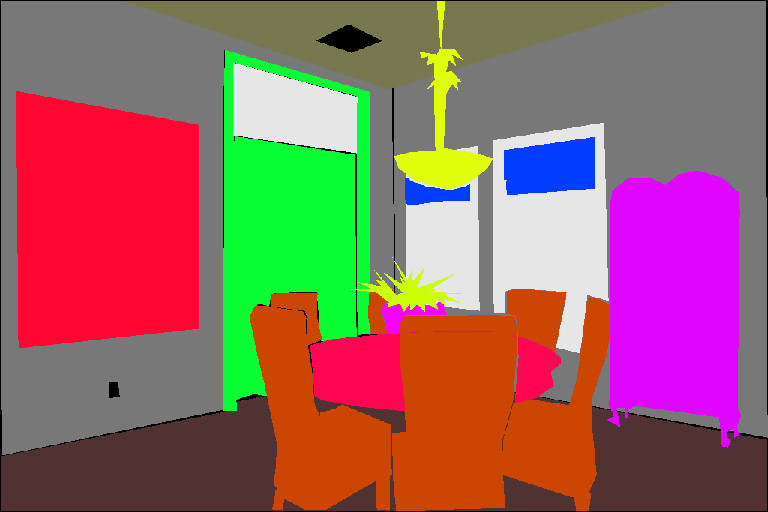}
\hspace{-12pt}
& \includegraphics[width=0.19\linewidth]{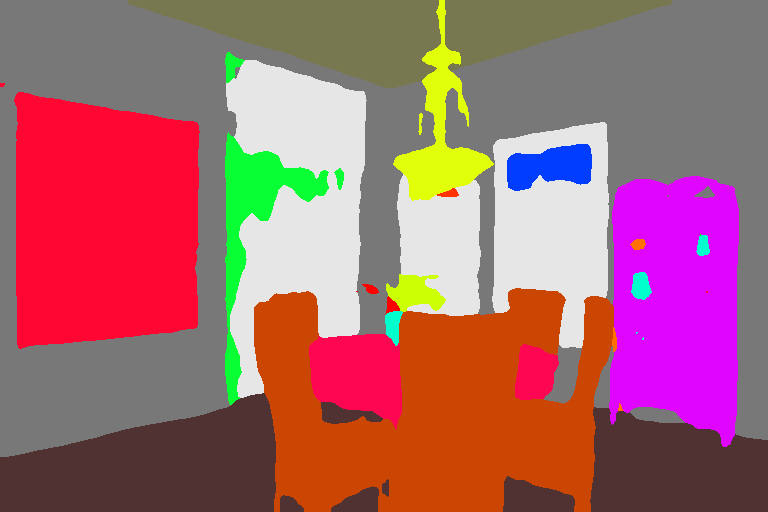}
\hspace{-12pt}
& \includegraphics[width=0.19\linewidth]{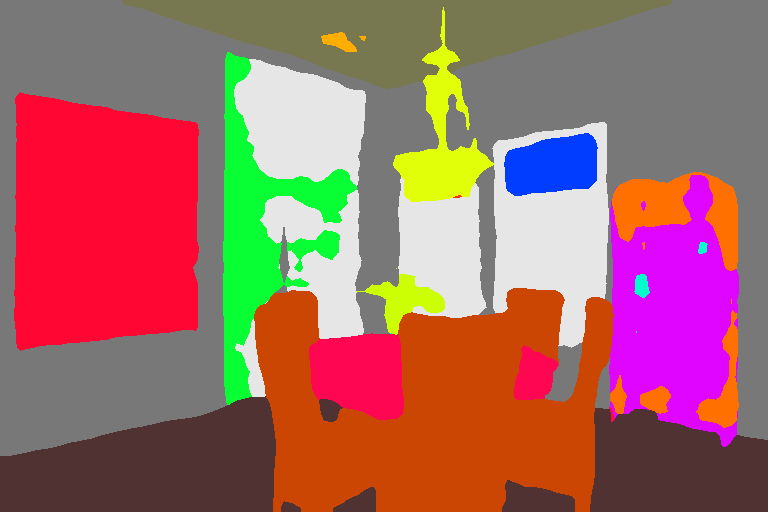}
\hspace{-12pt}
& \includegraphics[width=0.19\linewidth]{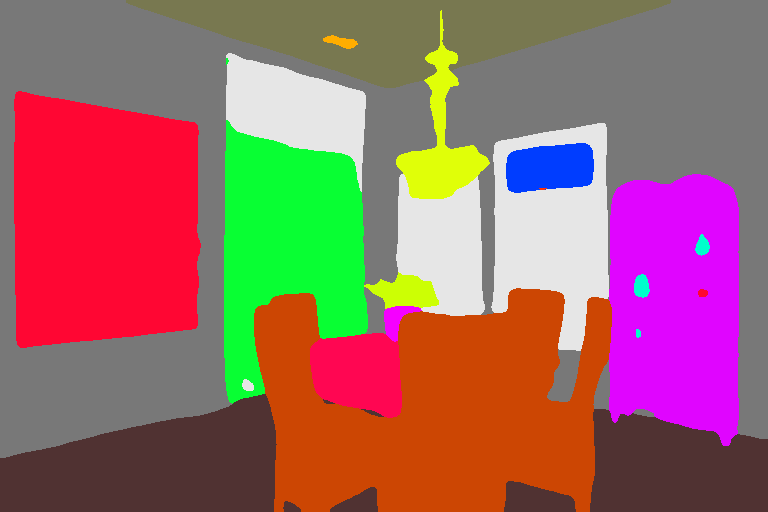}
\\

\includegraphics[width=0.19\linewidth]{images/ade/image/ADE_val_00001824.jpg}
\hspace{-12pt}
& \includegraphics[width=0.19\linewidth]{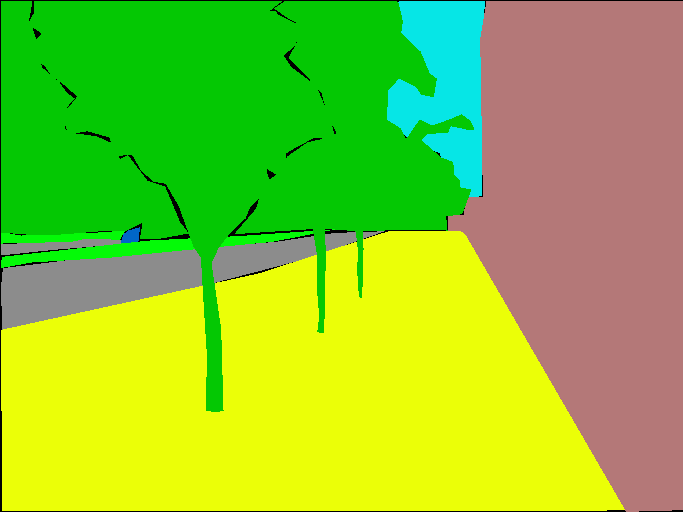}
\hspace{-12pt}
& \includegraphics[width=0.19\linewidth]{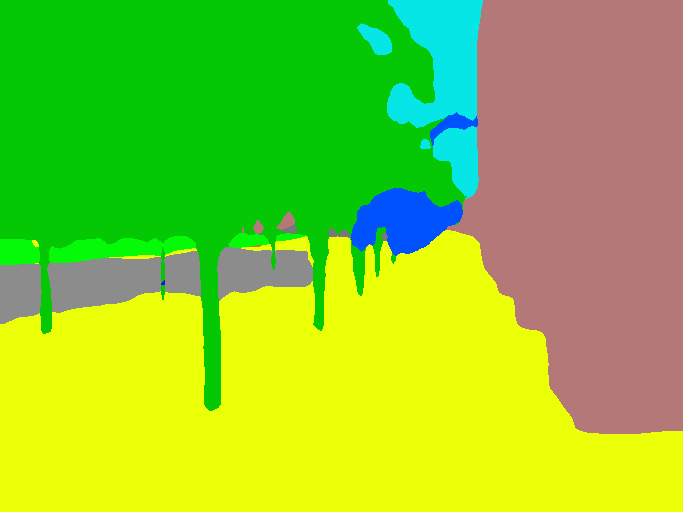}
\hspace{-12pt}
& \includegraphics[width=0.19\linewidth]{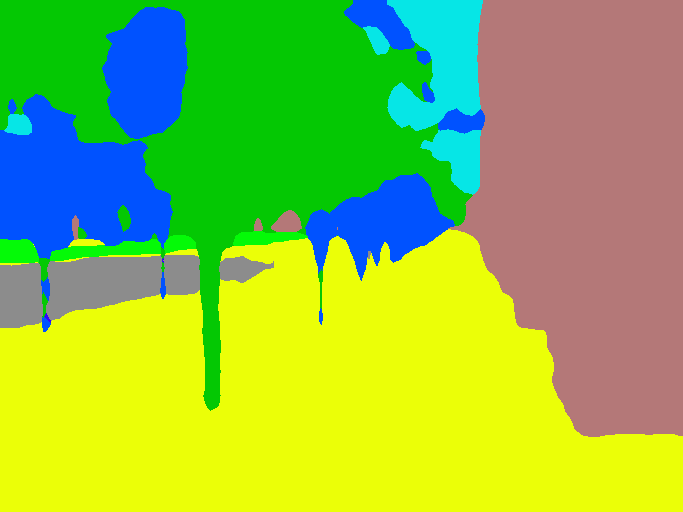}
\hspace{-12pt}
& \includegraphics[width=0.19\linewidth]{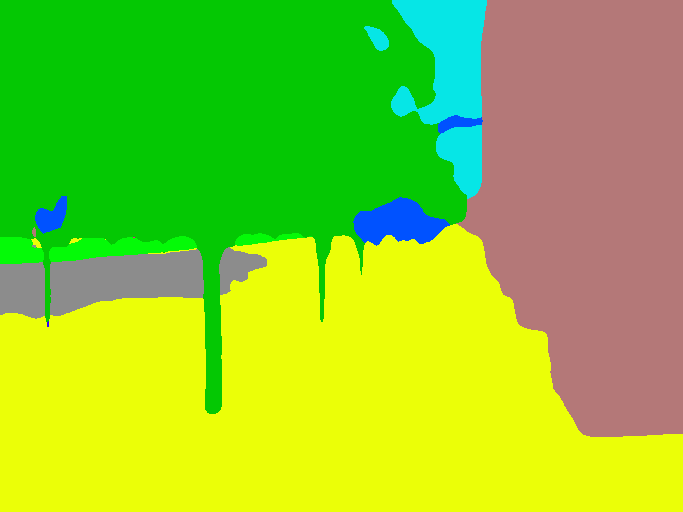}
\\

\includegraphics[width=0.19\linewidth]{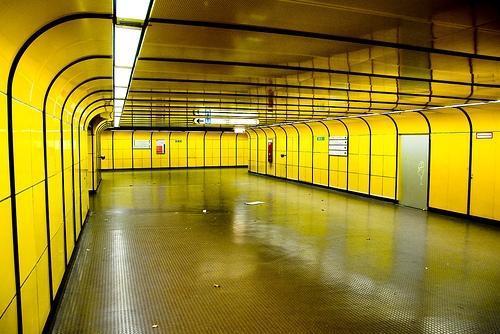}
\hspace{-12pt}
& \includegraphics[width=0.19\linewidth]{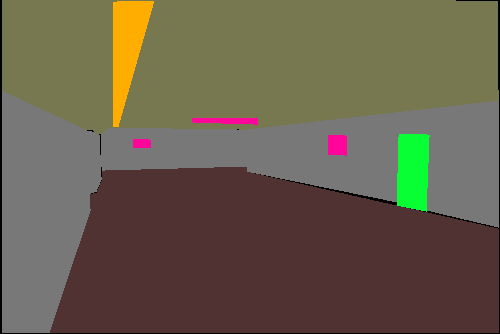}
\hspace{-12pt}
& \includegraphics[width=0.19\linewidth]{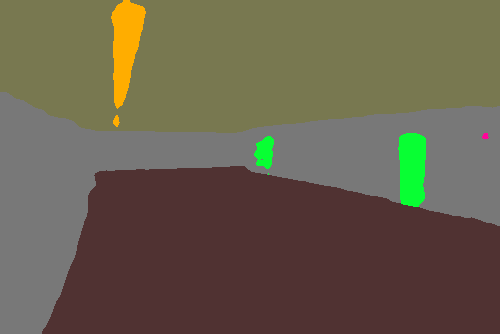}
\hspace{-12pt}
& \includegraphics[width=0.19\linewidth]{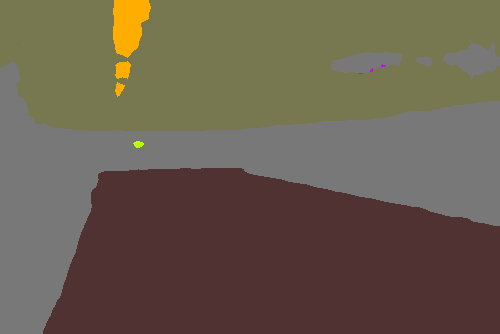}
\hspace{-12pt}
& \includegraphics[width=0.19\linewidth]{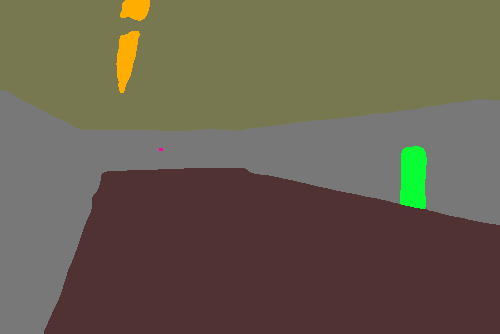}
\\

\includegraphics[width=0.19\linewidth]{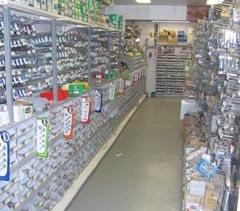}
\hspace{-12pt}
& \includegraphics[width=0.19\linewidth]{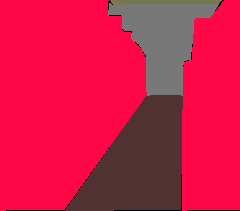}
\hspace{-12pt}
& \includegraphics[width=0.19\linewidth]{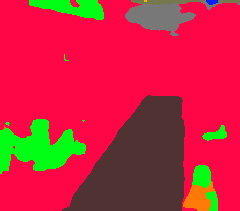}
\hspace{-12pt}
& \includegraphics[width=0.19\linewidth]{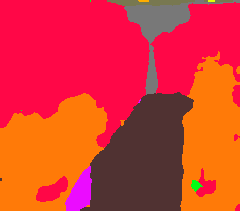}
\hspace{-12pt}
& \includegraphics[width=0.19\linewidth]{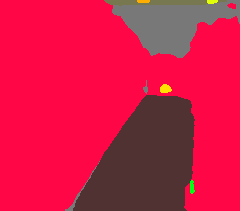}
\\

Image & Ground Truth & Unpruned & DCFP-60\% & SIRFP-60\%~(Ours) \\
\end{tabular}
\end{center}
\caption{
Visualization comparison with DCFP~\cite{dcfp} on ADE20K validation set with 60\% FLOPs reduction of Deeplabv3-ResNet50. Black in GT denotes the pixels which should be ignored for evaluation.
}
\label{fig:ade}
\end{figure*}

To compare the quality of segmentation results, we provide the visualization results in Figure~\ref{fig:city} and Figure~\ref{fig:ade} down below.
Compared with the prediction of DCFP~\cite{dcfp}, our results are more accurate on the details in the images, such as the thin poles.
The proposed SIRFP also performs better than DCFP on large objects and stuff such as the truck and sidewalks on Cityscapes.
On ADE20K, the performance of SIRFP exceeds DCFP on objects with both simple and complex shapes such as windows, doors, bridges, and trees.
Sometimes our results are even better than those of the unpruned original model, as shown in the second images in Figure~\ref{fig:city} and images in almost all the images except the fourth one in Figure~\ref{fig:ade}.
These demonstrate that SIRFP can retain or even enhance the spatial details in the prediction, while still preserving accurate high-level semantic information.

\section{C.~More Ablation Studies}
In this section, we provide more ablation experiments to help our readers better understand the rationale of our proposed redundancy metric.

\begin{table}[t!]
  \caption{Ablation of different feature resolutions on Cityscapes dataset using Deeplabv3-ResNet50. The resolution represents the size of features used for the redundancy calculation. }
  \label{tab:resolution}
  \centering
  \small
    \resizebox{\linewidth}{!}{
    \begin{tabular}{lccc}
    \toprule
     \textbf{Method}  & \bf \makecell{FLOPs  Reduction} & \textbf{Resolution} & \textbf{mIoU}(\%) \\
    \midrule  
     Unpruned &  0\%  & - & 79.3 \\
     \midrule
     \smallskip
      SIRFP-vector & 60\% & $1\times1$ & 80.4 \\
      \smallskip
      SIRFP-quarter & 60\% & $\frac{H}{4}\times \frac{W}{4}$ & 80.6 \\
      \smallskip
      SIRFP-half & 60\% & $\frac{H}{2}\times \frac{W}{2}$ & 80.8 \\
      \smallskip
    \bf SIRFP-full~(Ours) & 60\% & $H\times W$ & \bf 81.3 \\
      SIRFP-twice & 60\% & $(2H)\times (2W)$ & \bf 81.3 \\
    \bottomrule
    \end{tabular}}

\end{table} 

\subsection{Ablation study on feature resolution}

The richness of spatial details is closely related to the resolution of feature maps.
As a spatial-aware redundancy metric, we studied the effectiveness of SIRFP based on different resolutions.
The results in Table~\ref{tab:resolution} show that a larger resolution will lead to a more powerful pruning result under the same FLOPs reduction ratio.
This consolidates our claim that rich spatial information plays a significant role in the pruning of segmentation networks.
However, raising the resolution does not necessarily further improve the mIoU results, comparing results of the SIRFP-full and SIRFP-twice.
As the computational costs also increase with the growth of the feature resolution, we finally adopt the full-sized feature maps for redundancy calculation.

\begin{table}[t!]
  \caption{Ablation of different feature resolutions on Cityscapes dataset using Deeplabv3-ResNet50. The resolution represents the size of features used for the redundancy calculation. }
  \label{tab:metric}
  \centering
  \small
    \resizebox{\linewidth}{!}{
    \begin{tabular}{lccc}
    \toprule
     \textbf{Method}  & \bf \makecell{FLOPs  Reduction} & \textbf{Metric} & \textbf{mIoU}(\%) \\
    \midrule  
     Unpruned &  0\%  & - & 79.3 \\
     \midrule
      SIRFP-Dot & 60\% & dot production & 80.5 \\
      SIRFP-KL & 60\% & KL divergence & 81.1 \\
      SIRFP-Dice & 60\% & dice score & 81.0 \\
    \bf SIRFP~(Ours) & 60\% & JS divergence & \bf 81.3 \\
    \bottomrule
    \end{tabular}}

\end{table} 

\subsection{Ablation study on the metric}

We conduct experiments to choose the better metric for our edge-weight calculation.
In the main text, we introduced that our method employs the Jensen-Shannon~(JS) divergence for edge-weight calculation, as shown in Equation~3.
We also provide the results of the dot product of flattened features, the dice score of features, and the Kullback-Leibler~(KL) divergence of features in Table~\ref{tab:metric}, which are denoted as SIRFP-Dot, SIRFP-Dice, and SIRFP-KL, respectively.
For SIRFP-Dice, SIRFP-KL, and the SIRFP, the feature is normalized as a probability score map for the calculation of these metrics.
The results demonstrate that our method performs well using the dice score, KL divergence, and JS divergence.
The significant performance gap between SIRFP-Dot with the others can be attributed to the fact that the dot production only becomes large when the corresponding elements of two features are both positive or negative at the same time.
This may result in the undesired phenomenon: when two features are identical in terms of absolute values of each element but completely reverse in terms of the signs of those values, the dot production of these two features is negative; when two features are completely identical in terms of the signs of each element but significantly diverse
in terms of the absolute value, the value of dot production is positive, which is still bigger than that of the aforementioned case.
Given that we want to quantify the redundancy of the spatial distribution and semantic information regardless of the overall difference in the signs, we adopt the JS divergence from the other three instead of the dot production for the best mIoU results. 

\begin{table*}[h!]
  \caption{Pruning and training settings on different segmentation datasets.}
  \label{tab:settings}
  \centering
  \small
    \resizebox{0.7\linewidth}{!}{
    \begin{tabular}{clccc}
    \toprule
     & \textbf{Parameters}  & \bf \makecell{Cityscapes} & \textbf{ADE20K} & \textbf{COCO stuff-10k} \\
    \midrule  
    \multirow{8}{*}{\bf \rotatebox[origin=c]{90}{Pre-train}} 
    & weight\_decay & 0.0005 & 0.0001 & 0.0001 \\
    & optimizer & SGD & SGD & SGD \\
    & batch\_size & 8 & 16 & 16 \\
    & iterations~($T^{pre}_{total}$) & 4000 & 16000 & 6000 \\
    & learning\_rate & 0.01 & 0.01 & 0.001 \\
    & aux\_loss\_weight & 0.4 & 0.4 & 0.4 \\
    & aux\_loss\_stage & 3rd & 3rd & 3rd \\
    & training\_image\_size & $769\times769$ & $512\times512$ & $512\times512$ \\
    \midrule 
    \multirow{6}{*}{\bf \rotatebox[origin=c]{90}{Prune}}  & $T_{step}$ & 2 & 1 & 2 \\
    & FLOPs\_reduction~(60\%) & \{0.30, 0.60\} & \{0.60\} & \{0.30, 0.60\} \\
    & FLOPs\_reduction~(70\%) & \{0.35, 0.70\} & \{0.70\} & \{0.35, 0.70\} \\
    & FLOPs\_reduction~(80\%) & \{0.40, 0.80\} & \{0.80\} & \{0.40, 0.80\} \\
    & $\alpha$ & 0.99 & 0.99 & 0.99 \\
    & max\_channel\_sparsity & 0.9 & 0.9 & 0.9 \\
    \midrule  
    \multirow{8}{*}{\bf \rotatebox[origin=c]{90}{Fine-tune}} 
    & weight\_decay & 0.001 & 0.0005 & 0.0001 \\
    & optimizer & SGD & SGD & SGD \\
    & batch\_size & 8 & 16 & 16 \\
    & iterations~($T^{fine}_{total}$) & \{4000, 36000\} & \{144000\} & \{6000, 54000\} \\
    & learning\_rate & 0.01 & 0.01 & 0.001 \\
    & aux\_loss\_weight & 0.4 & 0.4 & 0.4 \\
    & aux\_loss\_stage & 3rd & 3rd & 3rd \\
    & training\_image\_size & $769\times769$ & $512\times512$ & $512\times512$ \\
    \bottomrule
    \end{tabular}
    }

\end{table*}

\section{D.~Implementation details}

Before we introduce the detailed settings, we have to point out that different segmentation networks share the same setting on segmentation datasets.
We provide the detailed pre-training, pruning, and fine-tuning settings in Talbe~\ref{tab:settings}.

Here, we also introduce the general pruning details on semantic segmentation. 
All of our segmentation experiments are conducted on NVIDIA RTX 3090 GPUs.
We adopt a global pruning technique to adaptively decide the channel sparsity of each layer.
To achieve this, we first set the pruning number of channels to $C^l-1$ for each layer to get the summation of edge weight $s_k$ of each $k$ as its importance score.
During each iteration in the EHGP~(Algorithm~1), we record and collect $s_k$.
Then we get the global threshold $s_th$ as the pruning decision indicator, such that pruning all the channels in the whole network whose $s_k\geq s_{th}$ satisfies the FLOPs reduction ratio.
Hence, the channel sparsity (or the number of channels to be pruned) in each layer can be decided by this threshold by setting the number of channels to be pruned as the number of channels whose $s_k\geq s_{th}$ in each layer.

For COCO2017, we set the learning rate, momentum, weight decay, batch of size, and training epochs as 0.0026, 0.9, 0.0005, 64, and 650, respectively.
We also adopt the `step' learning rate schedule. The learning rate is multiplied by 0.1 at the 430th and 540th epochs.
During training, the network is pruned once for every 2 epochs, from the 10th epoch to the 470th epoch.

On ImageNet, we set the learning rate, momentum, weight decay, batch size, and training epochs as 1.024, 0.875, 0.00003, 256, and 250, respectively.
We also adopt the `cosine' learning rate schedule. 
During training, the network is pruned once for every 2 epochs, from the 2-th epoch to the 180-th epoch.

\section{E.~Broader Impacts}
As technology advances, there is a growing demand for deploying segmentation neural networks on mobile devices.
Powerful semantic segmentation networks often require more computational resources, consume additional time, and incur higher power consumption during inference. Network pruning emerges as an effective strategy to enhance inference speed and conserve computational resources. Pruned networks demonstrate improved efficiency when executing complex segmentation tasks on mobile devices such as smartphones, smart vehicles, and wearable gadgets.

\end{document}